\documentclass{article}

\usepackage{microtype}
\usepackage{graphicx}
\usepackage{subcaption}
\usepackage{booktabs} % for professional tables

\usepackage{hyperref}

\usepackage[preprint]{icml2026}

\usepackage{amsmath}
\usepackage{amssymb}
\usepackage{mathtools}
\usepackage{amsthm}

\usepackage[capitalize,noabbrev]{cleveref}

\usepackage{amsmath,amsfonts,bm}

\def\eqref#1{equation~\ref{#1}}
\def\1{\bm{1}}

\DeclareMathAlphabet{\mathsfit}{\encodingdefault}{\sfdefault}{m}{sl}
\SetMathAlphabet{\mathsfit}{bold}{\encodingdefault}{\sfdefault}{bx}{n}

\usepackage{url}
\usepackage{algorithm}
\usepackage{multirow}   % for \multirow
\usepackage[table]{xcolor}
\usepackage{siunitx}
\usepackage{threeparttable}
\usepackage{enumitem}
\usepackage{float}
\usepackage{wrapfig}

\usepackage{algpseudocode}    % Now load the new package cleanly

\definecolor{customblue}{RGB}{135,200,235} % light but not too light blue
\newcommand{\ours}[1]{\cellcolor{customblue!30}#1}
\newcommand{\others}[1]{\cellcolor{gray!10}#1}

\definecolor{codeblue}{RGB}{0,70,127}
\definecolor{codegreen}{RGB}{50,140,80}
\definecolor{codelightgray}{RGB}{245,245,245}
\definecolor{commentgreen}{RGB}{0,128,0}
\icmltitlerunning{SINQ: \underline{Si}nkhorn-\underline{N}ormalized \underline{Q}uantization }
\begin{document}

\twocolumn[
  \icmltitle{SINQ: \underline{Si}nkhorn-\underline{N}ormalized \underline{Q}uantization \\ for Calibration-Free Low-Precision  LLM Weights}




  \begin{icmlauthorlist}
    \icmlauthor{Lorenz K. Muller}{comp}
    \icmlauthor{Philippe Bich}{comp}
    \icmlauthor{Jiawei Zhuang}{comp}
    \icmlauthor{Ahmet Celik}{comp}
    \icmlauthor{Luca Benfenati}{comp}
    \icmlauthor{Lukas Cavigelli}{comp}
  \end{icmlauthorlist}

  \icmlaffiliation{comp}{Huawei}

  \icmlcorrespondingauthor{Lorenz K. Muller}{lorenz.mueller@huawei.com}

  \icmlkeywords{Machine Learning, ICML}

  \vskip 0.3in
]



\printAffiliationsAndNotice{}  % no special notice (required even if empty)

\begin{abstract}
Post-training quantization has emerged as the most widely used strategy for deploying large language models at low precision. Still, current methods show perplexity degradation at bit-widths $\leq 4$, partly because representing outliers causes precision issues in parameters that share the same scales as these outliers. This problem is especially pronounced for calibration-free, uniform quantization methods. We introduce SINQ to augment existing post-training quantizers with an additional second-axis scale factor and a fast Sinkhorn–Knopp–style algorithm that finds scales to normalize per-row and per-column variances. 
We show that this approximates activation-aware quantization by recovering column scales from the weight matrix structure that are predictive of the typical activation magnitudes the matrix received during training. 
Our method has no interactions between layers and can be trivially applied to new architectures to quantize any linear layer.
We evaluate our method on the Qwen3 model family, among others. SINQ reduces the perplexity gap on WikiText2 and C4 by over $50\%$ against uncalibrated uniform quantization baselines, incurs zero to negligible compute overhead, and can be further enhanced by combining it with calibration and non-uniform quantization levels. Code is available at \url{https://github.com/huawei-csl/SINQ}.
\end{abstract}

\section{Introduction}
\label{sec:intro}

Post-training quantization (PTQ) is a powerful approach to reducing the cost of neural network inference. Weight quantization reduces the storage, memory, and data movement required to run a neural network. As such, it is useful on its own whenever any of these components bottleneck an inference system's performance. 
%When integer (INT) or floating-point (FP) weight quantization is combined with INT or FP activation quantization, it can also be used to reduce compute requirements by executing MatMul operations at low precision. 
Potential speed-ups of pure weight-quantization are substantial: For example, moving from float16 to int4 weights yields a potential speedup of 4x in memory-bound scenarios. Weight-only quantization is especially popular in LLM deployment because accelerator memory capacity and data movement are often the initial performance bottlenecks in this scenario.

In this paper, we demonstrate that a carefully chosen uncalibrated, uniform quantizer can approach the end-to-end output quality of calibrated quantizers or non-uniform formats while being appreciably simpler: Calibration (and even more so end-to-end optimization) is an intuitive approach to improving the output quality of quantized models, but comes with the inherent downsides of possible bias and overfitting \cite{AWQ} and additional compute time required at quantization time (which is prohibitive in continual learning settings). Similarly, non-uniform formats can offer an improvement over integer quantization \cite{NF4}, but require potentially costly look-ups during inference and cannot be combined with activation quantization in compute-limited scenarios. In brief, if uncalibrated uniform quantization were to reach the same output quality, it would be preferable for these reasons. This paper largely closes the gap between these different approaches to quantization.

% ...
\begin{figure*}[t]
  \centering 
  \includegraphics[width=.95\linewidth]{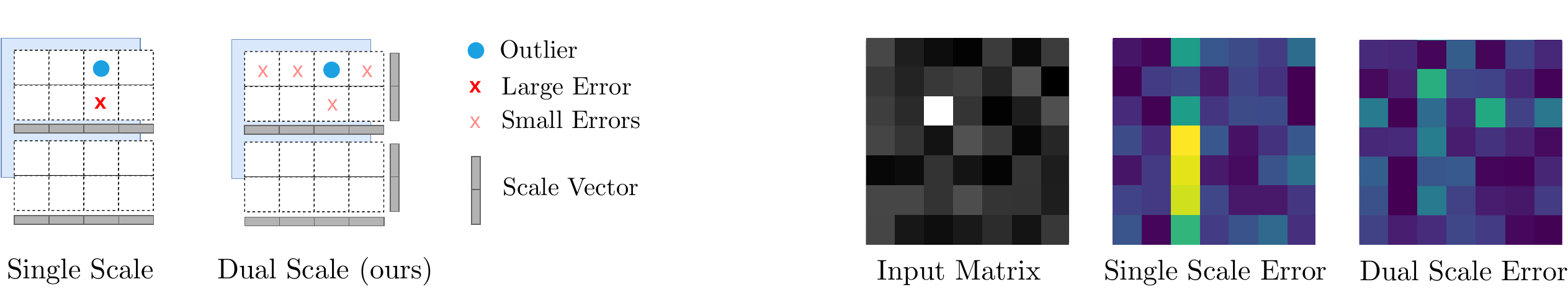}

  \caption{If we have scales along both dimensions of a matrix that is to be quantized, we can trade off the impact of outliers between rows and columns, which is impossible in single-scale quantization. Left: Conceptual illustration of error distributions with single or dual-scaling. Right: Example on small matrix.}
  \label{fig:dual_scale_idea}
\end{figure*}

The key contributions of this paper are:
\begin{itemize}[leftmargin=5mm]
   \item We find that column-wise weight matrix standard deviations are predictive of input scales, allowing for calibration-free pseudo-activation-aware quantization.
    \item We propose adding a scaling factor along the second axis of to-be-quantized matrix tiles alongside a fast algorithm based on Sinkhorn-Knopp iterations that finds column-wise scales that mimic activation-awareness, while avoiding row-wise kurtosis increase  (Sec.~\ref{sec:method}).
    % \item We demonstrate that the scales found by this method correlate with the typical input power to a given layer, enabling a calibration-data-free activation-aware quantization.
    \item In numerous experiments across different model scales, we show that our method substantially improves over state-of-the-art baselines for calibration-free quantization methods (Sec.~\ref{sec:experiments}).
    \item We provide code for easy quantization of LLMs using linear layers.
\end{itemize}

%PTQ can be divided into three categories that vary in the extent to which they optimize for a specific subset of the original model's training data: 1) Calibration-free approaches only consider the parameters of the original model; 2) calibrated methods compute statistics over some input data; and 3) end-to-end optimized methods use input data for gradient-based optimization.
%[How to position ourselves as calibration-free and calibrated, but not end-to-end optimized?]
\section{Methods}
\label{sec:method}
%To better understand the current landscape of post-training quantization methods, we identify four subdomains of model quantization: quantized parameterization, optimization methods, optimization spaces, and optimization targets. In the following, we separately consider possible improvements to these subdomains.
We divide our method into two parts: Firstly, the quantized parameterization, i.e. the mathematical expression used to map between the full precision and the quantized matrix. All quantization methods used in practice, have some set of auxiliary parameters to use in this mapping. Secondly, the representation space, i.e. the space in which we instantiate the full precision matrix when quantizing it. 

\subsection{Quantized Parametrization } 
Typically, one does not simply replace the weight matrix with, for example, an INT4 matrix, but rather divides it into tiles and assigns some higher-precision auxiliary parameters to each tile. Here, we describe different possibilities for the type of auxiliary parameters to use and how to tile the matrix.

\subsubsection{Parameterization per Tile}
% We consider uniform quantization (for non-uniform approaches see \ref{sec:related}).
\paragraph{Scales + Shifts}The most widely used approach uses a scale and a shift vector (e.g., \cite{HQQ}), like so:
\begin{equation}
\label{eq:zp}
    \mathbf{W}_\text{approx} = \vec{s} \odot (\mathbf{Q} + \vec{z} )
\end{equation}
where $\mathbf{W}_\text{approx}$ is a $N\times M$ matrix (or matrix tile), $\vec{s}$ is a $N\times 1$ vector, $\vec{z}$ is a $N \times 1$ vector and $\mathbf{Q}$ is a quantized $N \times M$ matrix. Also, the transpose of this with $1 \times M$ vectors is commonly used. 
All commonly used methods currently use this approach; specifically, all methods we compare in this paper do so.

\paragraph{Dual-Scales} Instead of supplying a single vector of scales along one dimension of the matrix, we supply two vectors, one along each dimension. Formulaically, we propose:
\begin{equation}
\label{eq:dual}
    \mathbf{W}_\text{approx} = \vec{s} \odot \mathbf{Q} \odot \vec{t} 
\end{equation}
where $\vec{s}$ is a $N\times 1$ vector, $\vec{t}$ is a $1 \times M$ vector and the rest is as above.

The key benefit of Eq.~\ref{eq:dual} can be illustrated as follows: Say $W_{ij}$ is an outlying large value. By scaling up $s_i$ and scaling down $t_j$ we can trade off quantization errors that will occur in row $i$ for errors in column $j$. See Fig.~\ref{fig:dual_scale_idea} for an illustration.

\paragraph{Dual-Scales + Shifts} If we do not mind the potential additional overhead (or rather, if an accuracy improvement justifies it), we can also add shifts to the dual scales:

\begin{equation}
\label{eq:dualshift}
    \mathbf{W}_\text{approx} = \vec{s} \odot (\mathbf{Q}+ \vec{z}) \odot \vec{t} 
\end{equation}

\subsubsection{Tiling}
Typically, tiling for quantization is implemented along one dimension of the matrix that is to be quantized (e.g., HQQ \cite{HQQ}, AWQ \cite{AWQ}). Consequently, these tiles have rectangular shapes; e.g., a $N \times M$ matrix tiled with tile size $T$ would yield tiles of shape $N \times T$. This could cause a problem with the dual-scale parameterization. Namely, the standard parameterization has $2 \times N \times M /T$ scale and shift parameters, while the dual-scaled only has $N \times M/T + M$. Because of this, we use dual-scale parameterization together with a shift (as in Eq.~\ref{eq:dualshift}). This has a small additional overhead compared to single-scale + shift parameterization; the total auxiliary parameters are $2 \times N \times M/T + M$. We report total memory use in all our experiments to confirm that this overhead of $M$ additional parameters is negligible.

\subsection{Representation Space} 
Before assigning values to the parameters from which we will reconstruct our matrix, we may want to transform the space in which the matrix is represented, to make the reconstruction better aligned with some quality metric (like weight MSE or end-to-end accuracy on some validation data).  The two most common among such transformations of the weight space are rotations (like the Hadamard transform \cite{quarot}), and channel-wise scaling (like in activation aware quantization (AWQ \cite{AWQ}) or Smoothquant \cite{smoothquant}). Here, we propose a new transformation of the weight matrix using our dual-scaling parameterization.

\subsubsection{Pseudo-Activation-Aware Quantization from Weight Structure}
\begin{figure*}
    \centering
        \begin{subfigure}{0.32\textwidth}
        \centering
        \includegraphics[width=\textwidth]{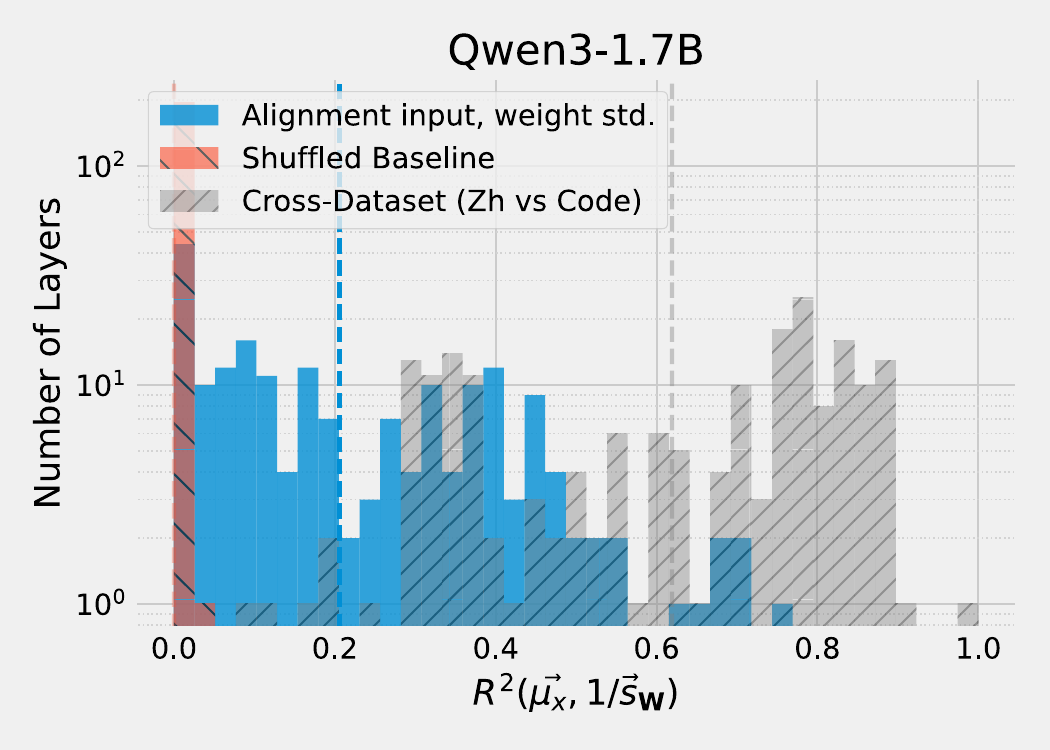}
        \caption{}
        \label{fig:corr}
    \end{subfigure}
    \hfill
    \begin{subfigure}{0.32\textwidth}
        \centering
        \includegraphics[width=\textwidth]{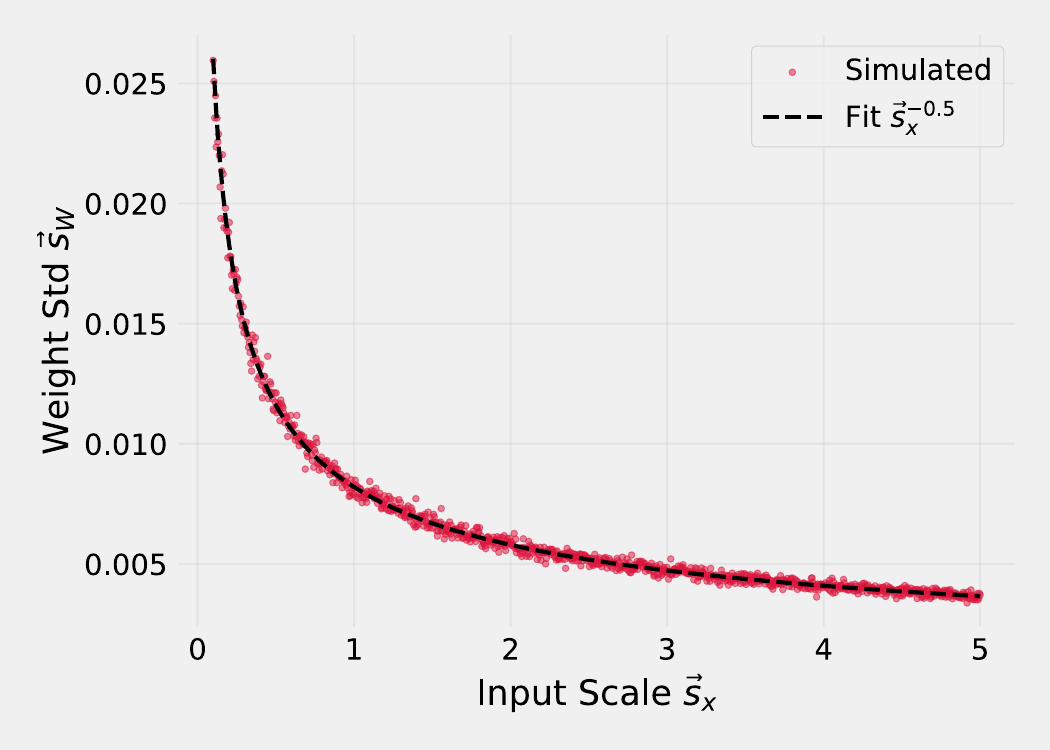}
        \caption{}
        \label{fig:adam}
    \end{subfigure}
    \hfill
    \begin{subfigure}{0.32\textwidth}
        \centering
        \includegraphics[width=\textwidth]{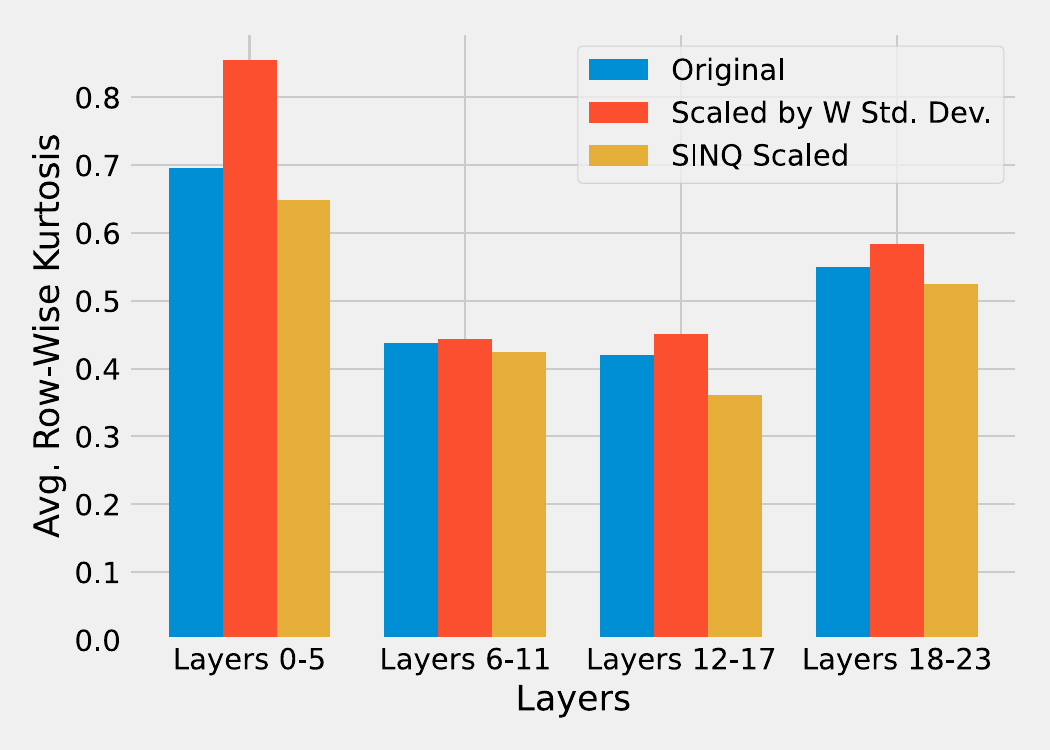}
        \caption{}
        \label{fig:kurtosis}
    \end{subfigure}
    \hfill
    \caption{
    (\subref{fig:corr}) In LLMs (here: Qwen3-1.7B) the reciprocal column-wise weight std. dev. is a good predictor of the average input magnitude (in terms of $R^2$-value), indicating a pseudo-activation-awareness. For comparison, the correlation when one vector is shuffled as a first baseline (that should be exceeded) and the cross-dataset correlation of average input magnitudes as a second baseline, Chinese wiki vs. Python code, as a plausible upper bound.
    (\subref{fig:adam}) In a simplified setting (a single layer model), Adam training yields a simple relation between weight std. dev. (per column) and input scales. The underlying reason is that weight updates depend on the outer product of incoming inputs and gradients.
    (\subref{fig:kurtosis}) Directly scaling weights with their column-wise std. dev. creates more outliers row-wise, evidenced by increased kurtosis. SINQ successfully avoids this increase, while still scaling columns with a correlate of input scales.
    }
    \label{fig:sinq_reason}
\end{figure*}

Prior work assumed that activation-awareness requires calibration data. Here, we find that analysis of the weight matrix can reveal information about the training data. Namely, we study the relationship between the per-column weight standard deviation $\vec{s}_\mathbf{W}$ and the sample mean of the absolute value of the input to a given layer, $\vec{\mu}_x$. We find that they have a strong predictive power of each other (see Fig.~\ref{fig:corr} and in appendix Fig.~\ref{fig:corr_combined}). We observe a similar relationship in all models we tested (incl. Qwen, Phi and Llama models). Notably, $\vec{\mu}_x$ is the quantity used in AWQ~\cite{AWQ} to determine pre-quantization column scales (see Sec.~\ref{sec:awq}). 

The underlying driver of this relation is that during training, weight updates are a function of the outer product of inputs and gradients. For simplicity, consider a single linear layer trained with input $\vec{x} \sim \mathcal{N}(\vec{0}, \vec{s}_x)$ using Adam~\cite{adam} on a noisy (Gaussian) target. In this setting, we observe (see Fig.~\ref{fig:adam})
\begin{equation}
    \vec{s}_\mathbf{W} \propto \frac{1}{\sqrt{|\vec{s}_x|}}.
\end{equation}

However, scaling each column by its inverse standard deviation directly is a suboptimal strategy for quantization. This naive scaling often worsens per-\emph{row}-outliers, which is a source of new quantization error (e.g. evidenced by an increased per-row kurtosis, see Fig.~\ref{fig:kurtosis}). Thus, we must balance two effects: normalize column standard deviations to approximate activation-aware quantization, while simultaneously keeping row standard deviations approximately constant to avoid creating a new source of quantization errors. To achieve this, we propose a Sinkhorn-Knopp-style algorithm to iteratively normalize row and column standard deviations (Alg.~\ref{alg:sinq}, further implementation details in code in supplementary).

This Algorithm is essentially a search to minimize the matrix imbalance $I$, which we define as
\begin{equation}
\label{eq:imba}
    I(\mathbf{W}) = \frac{\max \left( \max_i \sigma_i^{\text{row}}, \max_j \sigma_j^{\text{col}} \right)}
                         {\min \left( \min_i \sigma_i^{\text{row}}, \min_j \sigma_j^{\text{col}} \right)}
\end{equation}
\noindent where $\sigma_i^{\text{row}}$ and $\sigma_j^{\text{col}}$ denote the standard deviation of the $i$-th row and $j$-th column of $\mathbf{W}$, respectively.

Empirically, we find that this algorithm not only successfully balances row and column standard deviations, but that the input-side scales $\vec{t}$ it yields have a higher coefficient of determination with $\vec{\mu}_x$ than the pure standard deviation $\vec{s}_\mathbf{W}$. %(see Fig.~\ref{fig:correlation}).

\begin{figure}
    \centering
    \begin{subfigure}{0.22\textwidth}
        \centering
        \includegraphics[width=\textwidth]{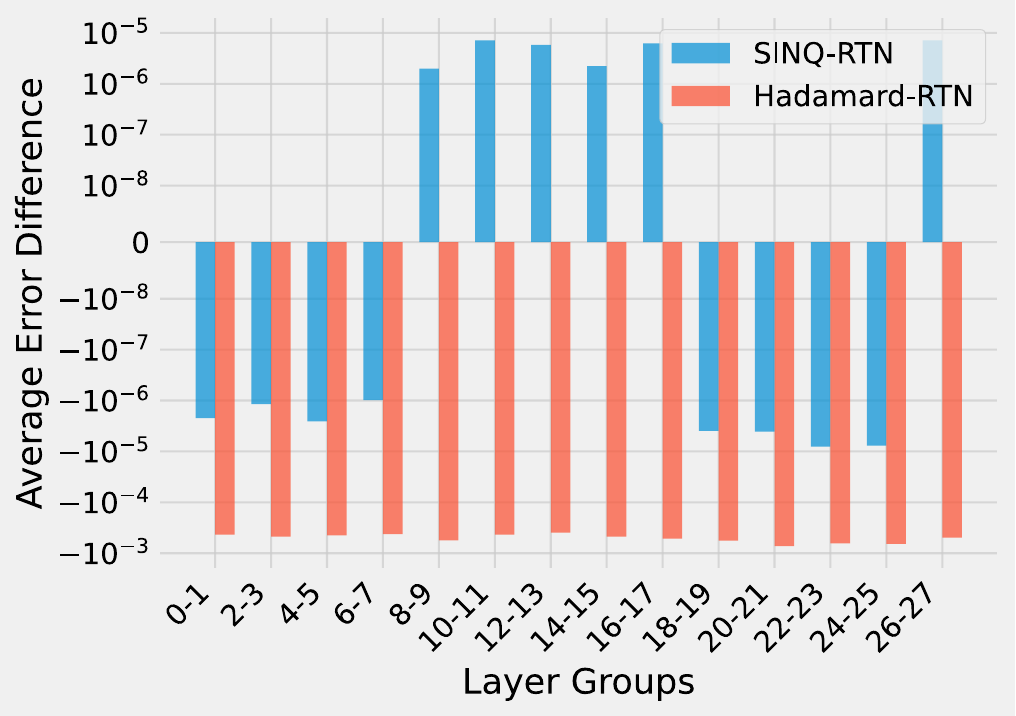}
        \caption{}
        \label{fig:mat_err}
    \end{subfigure}
    \hfill
    \begin{subfigure}{0.22\textwidth}
        \centering
        \includegraphics[width=\textwidth]{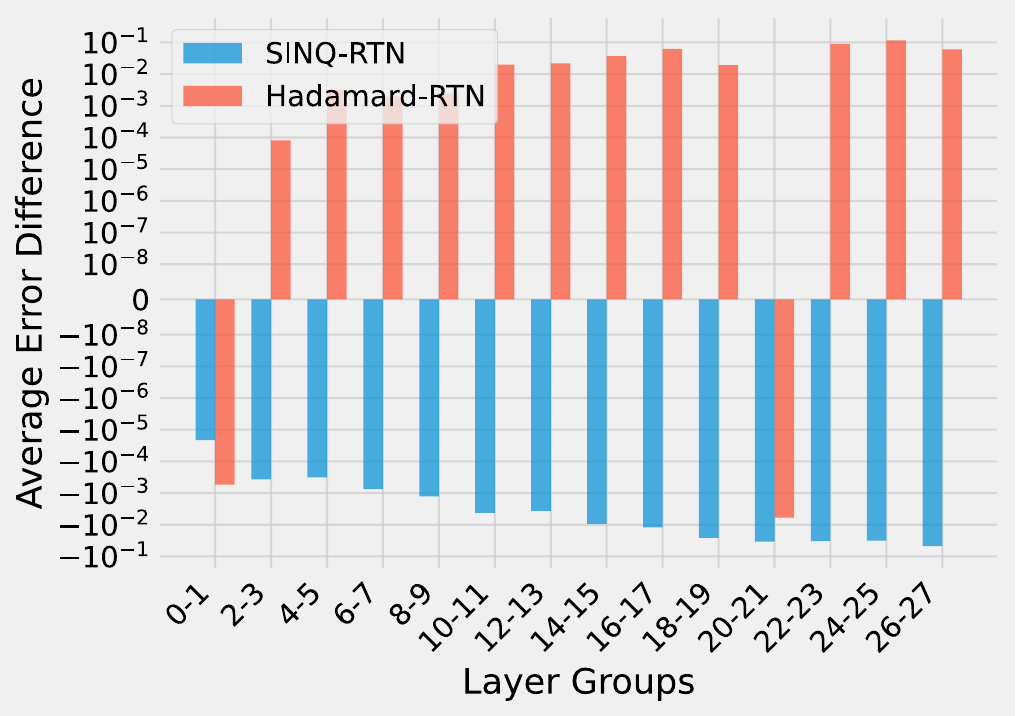}
        \caption{}
        \label{fig:act_err}
    \end{subfigure}
    \hfill
    \caption{
    (a) Matrix reconstruction error with SINQ and Hadamard rotation plus RTN compared to RTN; y-axis: $E_\text{i} - E_\text{rtn}$, where $i$ is either SINQ or Hadamard (negative values: $i$ improves over RTN, positive values: $i$ is worse than RTN). 
    (b) Activation reconstruction error in similar setting.
    In all cases Hadamard gives better matrix reconstruction, but SINQ better activation reconstruction in most cases. Evaluated on attention layers in Qwen3-1.7B. }
    \label{fig:sinq_vs_had}
\end{figure}

Further evidence that SINQ exhibits a calibration-free activation-awareness, is that it does not necessarily reduce the matrix reconstruction error, but it does reduce the activation reconstruction error of the output of the quantized layer. This stands in contrast to the standard uncalibrated space-transformation method: Hadamard rotation. In Fig.~\ref{fig:sinq_vs_had} we show precisely this: Hadamard rotation yields a better matrix reconstruction, but sometimes increases the output activation error over RTN.  With SINQ, it is the other way around.

\begin{algorithm}[t]
\small
\caption{SINQ: A dampened Log-Space Sinkhorn-inspired Normalization. Iteratively normalize the standard deviation of the rows and columns of the matrix to be quantized. Then apply a standard quantization method.}
\label{alg:sinq}
\begin{algorithmic}[1]
\Require Weight matrix $\mathbf{W} \in \mathbb{R}^{m \times n}$, Iterations $K$, Bits $b$, Step-Sizes $s_\text{min}, s_\text{max}$
\Ensure Quantized weights $\mathbf{Q}$, Scales $\vec{s} \in \mathbb{R}^m, \vec{t} \in \mathbb{R}^n$

% Initialization
\State $\vec{\sigma}_{\text{row}} \gets \vec{\sigma}^{\text{row}}(\mathbf{W}) ; \quad \vec{\sigma}_{\text{col}} \gets \vec{\sigma}^{\text{col}}(\mathbf{W})$
\State $\tau \gets \min(\min(\vec{\sigma}_{\text{row}}), \min(\vec{\sigma}_{\text{col}}))$ \Comment{Target variance}
\State $\mathbf{u} \gets \vec{0}_m ; \quad \mathbf{v} \gets \vec{0}_n$ \Comment{Initialize log-scales}
\State $\theta^* \gets \{ \mathbf{u}, \mathbf{v} \}; \quad I_{\text{best}} \gets \infty$ \Comment{Track best solution}

\For{$k \gets 1$ to $K$}
    \State $\mathbf{\hat{W}} \gets (\mathbf{W} \oslash \exp(\mathbf{u})) \oslash \exp(\mathbf{v})$ \Comment{Apply current scales}
    
    % Track best-so-far imbalance
    \State $I_{\text{curr}} \gets \text{Imbalance}(\mathbf{\hat{W}})$
    \If{$I_{\text{curr}} < I_{\text{best}}$}
        \State $I_{\text{best}} \gets I_{\text{curr}}; \quad \theta^* \gets \{ \mathbf{u}, \mathbf{v} \}$ \Comment{Snapshot best scales}
    \EndIf

    % Compute parallel damped updates in log-space
    \State $\vec{\delta}_{\text{col}} \gets \log \left( \text{clamp}(\vec{\sigma}^{\text{col}}(\mathbf{\hat{W}}) / \tau, ~s_\text{min}, ~s_\text{max}) \right)$
    \State $\vec{\delta}_{\text{row}} \gets \log \left( \text{clamp}(\vec{\sigma}^{\text{row}}(\mathbf{\hat{W}}) / \tau, ~s_\text{min}, ~s_\text{max}) \right)$%\footnote{the clamp function sets values in the first argument less than the second argument and greater than the third argument to those values respectively.}

    \State $\mathbf{v} \gets \mathbf{v} + \vec{\delta}_{\text{col}}$ \Comment{Update col log-scales}
    \State $\mathbf{u} \gets \mathbf{u} + \vec{\delta}_{\text{row}}$ \Comment{Update row log-scales}
\EndFor

% Final Recovery
\State $\vec{s}, \vec{t} \gets \exp(\mathbf{u}^*), \exp(\mathbf{v}^*)$ \Comment{Recover best linear scales}
\State $\mathbf{\hat{W}} \gets (\mathbf{W} \oslash \vec{s}) \oslash \vec{t}$
\State $\mathbf{Q}, \vec{z}, \vec{s}_{q} \gets \text{RoundToNearest}(\mathbf{\hat{W}}, b)$

\State \Return $\mathbf{Q}, \vec{z}, (\vec{s}_{q} \odot \vec{s}), \vec{t}$ 
\end{algorithmic}
\end{algorithm}

\subsubsection{Activation-aware Calibration with SINQ}
\label{sec:awq}
\label{sec:asinq}
Applying the proposed normalization of Alg.~\ref{alg:sinq} is also helpful in combination with calibration. Consider, for example, AWQ \cite{AWQ}.
%While the weight standard deviation is a good proxy for calibration data, there is still some residual benefit to observing input scales from a calibration set. For this we can combine AWQ with SINQ. In the setup we describe below, the awq-scales improve the weighting of matrix entries by importance, while Alg.~\ref{alg:sinq} contributes row-variance normalization. 

AWQ finds a vector of scales for each input of a linear layer by minimizing the 2-norm between the linear layer's output with the original and the scaled, quantized weight matrix. Formally,
\begin{equation}
    \alpha^* = \text{argmin}_\alpha \left\|  \vec{x}\cdot \mathbf{W}^\mathsf{T} - (\vec{x} \oslash \vec{\mu_x}^\alpha) \cdot d_q(q(\vec{\mu_x}^\alpha \odot \mathbf{W}))^\mathsf{T} \right\|_2 ,
\end{equation}
where $\vec{x}$ is a set of inputs, $\vec{\mu_x}$ is the sample mean of the absolute value of $\vec{x}$, $q(\cdot)$ is the quantization function, $d_q(\cdot)$ is the dequantization function and $\alpha^*$ is a per-layer parameter (a scalar).\footnote{For results in combination with our method, we modify this formula by changing the norm to a 1-norm, which we observe to give slightly better results in combination with SINQ.} The final scale used is $\vec{\mu_x}^{\alpha^*}$ (i.e. the minimizer of the activation reconstruction error). 

We find that normalizing first with SINQ in this setting helps prevent the row-wise kurtosis from increasing too much when applying AWQ scales (similar to Fig.~\ref{fig:kurtosis}), see Fig.~\ref{fig:awq-asinq-kurt}. We term the combination of SINQ with AWQ `ASINQ'.

\subsection{Implementation Considerations}
The second scale $\vec{t}$ can be applied as a scale vector to the input of the quantized linear layer, rather than when reconstructing the weight (see Eq.~\ref{eq:vector-grouping}). In this formulation, the forward complexity of the dual-scaling approach becomes very similar to AWQ: The term inside the square bracket is the RTN dequantization, and for each linear layer, we need to do one additional element-wise scaling of activations (just like in AWQ). 
\begin{eqnarray}
\label{eq:vector-grouping}
        \vec{x} \cdot \mathbf{W}_\text{approx}^\mathsf{T} &=& \vec{x} \cdot \left[\vec{s} \odot (\mathbf{Q}+ \vec{z}) \odot \vec{t} \right]^\mathsf{T} \nonumber \\
        &=& \left( \vec{x} \odot \vec{t} \right)  \cdot \left[\vec{s} \odot (\mathbf{Q}+ \vec{z})\right]^\mathsf{T}
\end{eqnarray}

The overhead of doing the additional scaling is small in practice, see Sec.~\ref{sec:inference-time}
.
\subsubsection{No-Overhead SINQ}
\label{sec:no-overhead}
To avoid the small overhead of additional element-wise scaling, we can absorb input scales into preceding layers. This comes with the caveat that for many commonly used models, some layers need to share this second scale. For example, in Qwen-3 models, the Q, K, and V-layers share an input scale and the Gate and Up-Projection share an input scale. In the experiments, we show that this implies a trade-off between output quality (Appendix~\ref{app:nooverhead}) and a minor inference time overhead (see Sec.~\ref{sec:inference-time}). No-overhead SINQ can also be used as a pure pre-processing step to improve the performance of other post-training quantization methods. E.g., for GGUF, we see improved accuracy while preserving end-to-end inference speed-up over fp16, see Appendix~\ref{app:gguf}.

\section{Experiments}
\label{sec:experiments}
\begin{table*}[t!]
\centering
\small
\setlength{\tabcolsep}{3.6pt}
\begin{threeparttable}
\caption{Weight-only uncalibrated uniform PTQ on Qwen3 models with 3-bit and 4-bit quantization, reporting perplexity and actual memory usage (GB). Lower is better for all metrics.  The best result for a given setting is marked in \textbf{bold}. }
\label{tab:qwen-ptq}
\begin{tabular}{l l l*{9}{r}}
\toprule
& & & \multicolumn{3}{c}{\textbf{Qwen3-1.7B}} & \multicolumn{3}{c}{\textbf{Qwen3-14B}} & \multicolumn{3}{c}{\textbf{Qwen3-32B}} \\
\cmidrule(lr){4-6} \cmidrule(lr){7-9} \cmidrule(lr){10-12}
& & Method 
& Mem. & \others{\textit{Wiki2}\,$\downarrow$} & \others{\textit{C4}\,$\downarrow$}
& Mem. & \others{\textit{Wiki2}\,$\downarrow$} & \others{\textit{C4}\,$\downarrow$}
& Mem. & \others{\textit{Wiki2}\,$\downarrow$} & \others{\textit{C4}\,$\downarrow$} \\
\midrule

% Baseline (FP16)
& & Original (BF16) & 3.44 & \others{\textit{16.67}} & \others{\textit{19.21}} & 29.54 & \others{\textit{8.64}} & \others{\textit{12.01}} & 65.52 & \others{\textit{7.60}} & \others{\textit{10.77}} \\
\midrule

% Calibration-free
% \multirow{10}{*}{\rotatebox[origin=c]{90}{\scriptsize\textsc{Calibration free}}}
& \multirow{4}{*}{\rotatebox[origin=c]{90}{\scriptsize\textsc{3-bit}}}
   & RTN\tnote{$\dagger$}                    & 1.28 & \others{32.43} & \others{31.10} & 9.23 & \others{10.50} & \others{14.88} & 17.61 & \others{30.78} & \others{35.83} \\
&  & Hadamard + RTN\tnote{$\dagger$}         & 1.28 & \others{32.40} & \others{31.07} & 9.23 & \others{10.60} & \others{15.10} & 17.61 & \others{11.26} & \others{14.83} \\
& & HQQ                 & 1.28 & \others{32.10} & \others{30.54} & 9.23 & \others{10.73} & \others{14.39} & 17.62 & \others{9.09} & \others{12.58} \\
& & \ours{\textbf{SINQ} \textit{(ours)}} & \ours{1.28} & \ours{ \textbf{22.39}} & \ours{ \textbf{24.88}} & \ours{9.25} & \ours{ \textbf{9.33}} & \ours{ \textbf{12.90}} & \ours{17.61} & \ours{\textbf{8.79}} & \ours{\textbf{11.83}} \\
\cmidrule(lr){2-12}
& \multirow{4}{*}{\rotatebox[origin=c]{90}{\scriptsize\textsc{4-bit}}}
 & RTN\tnote{$\dagger$}                & 1.42 & \others{18.74} & \others{20.81} & 10.54 & \others{8.95} & \others{12.50} & 20.78 & \others{8.92} & \others{12.80} \\
%& & BnB (FP4)             & 1.42 & \others{24.05} & \others{23.44} & 10.59 & \others{8.88         } & \others{12.54} & 20.67 & \others{11.93} & \others{16.90} \\
%& & BnB(NF4)             & 1.42 & \others{18.00} & \others{20.43} & 10.59 & \others{8.89         } & \others{12.27} & 20.67 & \others{7.94} & \others{11.21} \\
%& & HIGGS (non-uniform)             & 1.51 & \others{23.98} & \others{25.27} & 10.28 & \others{9.13        } & \others{12.56} & 19.88 & \others{8.02} & \others{11.24} \\
& & Hadamard + RTN\tnote{$\dagger$}      & 1.42 & \others{19.10} & \others{20.70} & 10.54 & \others{8.85} & \others{12.35} & 20.78 & \others{8.28} & \others{11.60} \\
& & HQQ             & 1.42 & \others{18.96} & \others{22.10} & 10.54 & \others{8.78} & \others{12.36} & 20.78 & \others{8.62} & \others{12.20} \\
& & \ours{\textbf{SINQ} \textit{(ours)}} & \ours{1.42} & \ours{\textbf{17.14}} & \ours{\textbf{ 19.83}} & \ours{10.56} & \ours{\textbf{8.76}} & \ours{\textbf{12.21}} & \ours{20.73} & \ours{\textbf{7.74}} & \ours{\textbf{10.96}} \\
% & & \ours{\textbf{SINQ no overhead} \textit{(ours)}} & \ours{1.42} & \ours{{17.63}} & \ours{{ 19.99}} & \ours{10.56} & \ours{{8.78}} & \ours{{12.32}} & \ours{20.73} & \ours{{7.78}} & \ours{{11.15}} \\
%& & \ours{\textbf{SINQ} (NF4) \textit{(ours)}} & \ours{1.42} & \ours{16.94} & \ours{19.83} & \ours{10.56} & \ours{8.72} & \ours{12.13} & \ours{20.73} & \ours{7.83} & \ours{10.97} \\
\bottomrule
\bottomrule
\end{tabular}
\begin{tablenotes}
  \item[$\dagger$] Baseline result obtained by running our own implementations.
\end{tablenotes}
\end{threeparttable}
\end{table*}

We evaluate our proposed methods against several strong baselines in 4-bit (and to a lesser extent 3-bit) quantization using the permissively licensed and powerful Qwen3 family of models by \cite{qwen3}. We use the evaluation settings of \cite{qwenquant}. In accordance with \cite{notaccuracy}, we report perplexities for language modeling and flip percentages for QA tasks. Flip percentages indicate how often the quantized model predicts a different result from the original full-precision model. Additionally, benchmark results for reasoning benchmarks are provided in the appendix. Code to reproduce the perplexities reported for our methods in this section is available in the supplementary. %Wherever possible we use the original code of other methods. In the case of 3-bit AWQ we use our own implementation (marked by *), because 3-bit quantization is not supported in auto-awq.

We highlight that SINQ is architecture agnostic, i.e., there is no interdependency between the quantization of different layers (unlike, e.g., in methods using Hadamard transformations). For all models we tried, it works out of the box.  Wherever there is no mention to the contrary, we set the group size to 64, batch-size to 8, and for SINQ use dual-scaling + shift parameterization.

To fairly account for the overhead of different parameterizations and tiling strategies, we report total memory use (including activations) in our experiments. %and look for Pareto-optimal parameterizations in the output quality vs. memory trade-off. 
%{\color{red} wikitext-2 and c4 citations?}

% {\color{red} Add table that lists 1) Uniform 2) Calibrated 3) Model Agnostic 4) output quality} 

\subsection{Uncalibrated Uniform Quantization}

\begin{figure*}[h]   % [t] -> top of column (ICLR likes top floats)
  \centering
  % 0.23\textwidth leaves ~2 % for separation; tweak for exact fit
  \hfill
  \subfloat[]{\includegraphics[width=0.45\textwidth]{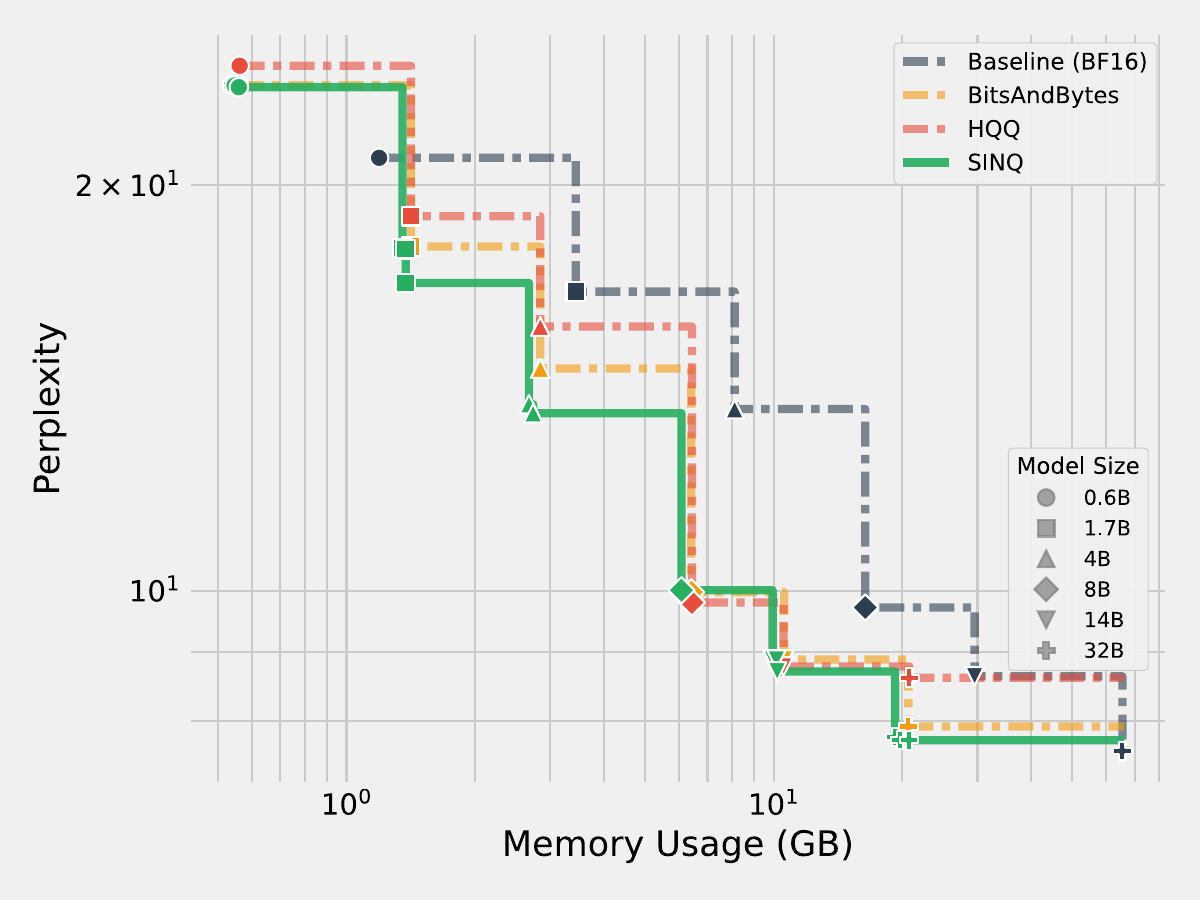}\label{fig:pareto4}}
  \hfill              % equal spacing between sub-floats
  \subfloat[]{\includegraphics[width=0.45\textwidth]{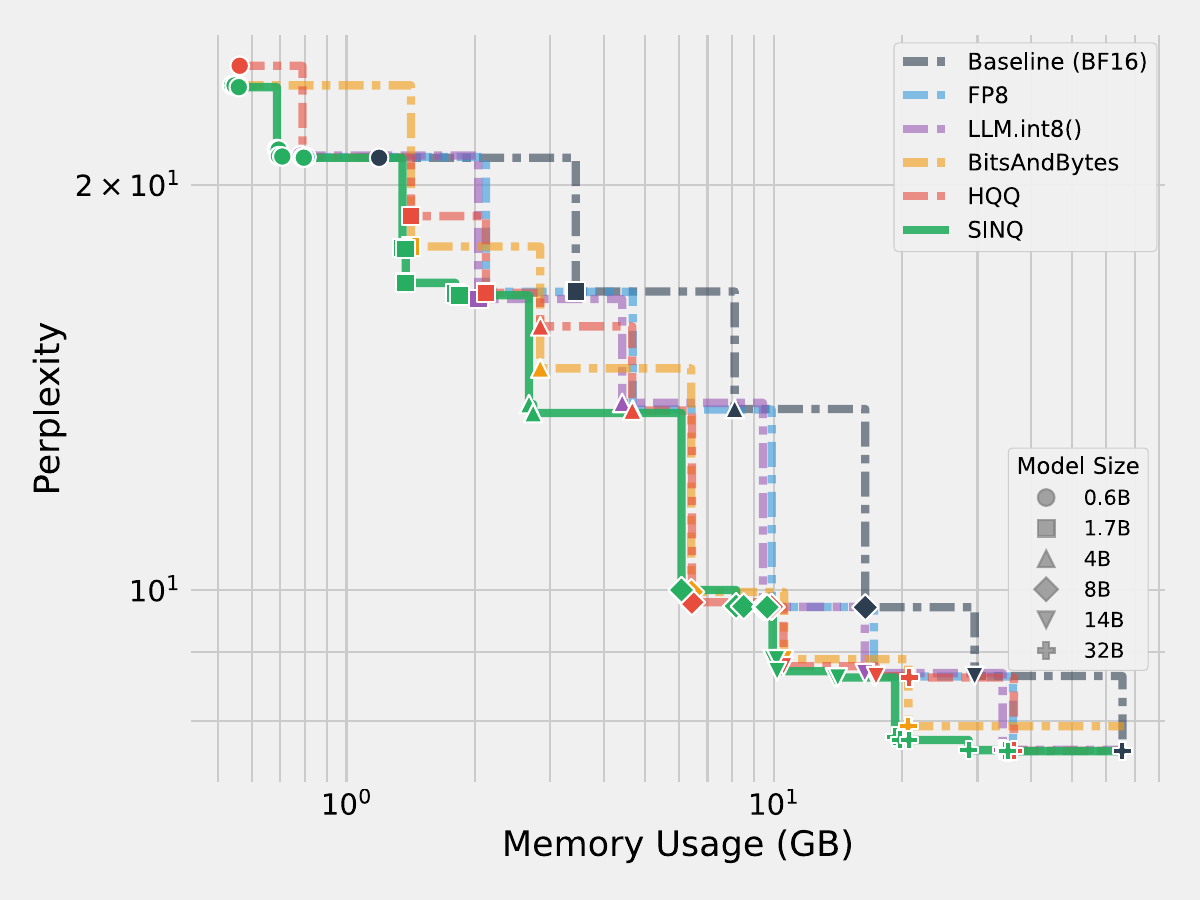}\label{fig:pareto468}}
  \hfill
  
  \caption{Pareto plot in terms of memory vs. WikiText2 perplexity for Qwen3-0.6B to 32B for different uncalibrated quantization methods. (a) compares different 4-bit methods (including FP4, INT4, and NF4 where available). The maximum distance from the 4-bit pareto front of our method is $<0.01$ppl. Note that the difference to the baseline is small. (b) allows bit widths of 4, 6, 8. For 8-bit quantization we inlcude \texttt{LLM.int8()} from~\cite{llmint8} as a reference method. Both plots include the BF16 model as a baseline. For these plots we allow group sizes 64 and 128 for all methods. }
  \label{fig:pareto_res}
\end{figure*}

In Tab.~\ref{tab:qwen-ptq}, %, we compare our method against various baselines. 
our method outperforms the baselines in every uncalibrated case in terms of C4 \cite{c4} and WikiText2 perplexity, sometimes reducing the residual difference to the 16-bit baseline by more than half. Similarly, our method performs best in terms of the average number of flips (see Tab.~\ref{tab:qwen-flips}). 
Fig.~\ref{fig:pareto_res} shows the memory-perplexity Pareto plot for different quantization methods across a wide range of Qwen3 models. Because the Qwen3 models are available in many different sizes, our method can dominate the bfloat16 baselines across a large range of available memory, from ca. 1.5 GB to 65 GB. Additional perplexity results on Llama (Sec.~\ref{sec:llama}), DeepSeek-V3 (Sec.~\ref{sec:app:dsv3}), Phi (Sec.~\ref{sec:phi}) and other Mixture-of-Experts (MoE, \cite{MoE}) models (Sec.~\ref{sec:app:moe}) can be found in the Appendix.

\begin{table*}[t]
\centering
\small
\setlength{\tabcolsep}{4pt}
\begin{threeparttable}
\caption{Flip rates (\%) (as proposed by \cite{notaccuracy} since more reliable than accuracy) on HellaSwag, PIQA, and MMLU for Qwen3 models with 3-bit and 4-bit quantization. Lower is better. The best result for a given setting is marked in \textbf{bold}. Accuracy measurements are given in the appendix.}
\label{tab:qwen-flips}
% \sisetup{detect-weight=true, detect-family=true, table-number-alignment=center, table-format=2.2} % adjust as needed
\begin{tabular}{l l l*{8}{r}}
\toprule
& & & \multicolumn{4}{c}{{\bfseries Qwen3-14B}} & \multicolumn{4}{c}{{\bfseries Qwen3-32B}} \\
\cmidrule(lr){4-7} \cmidrule(lr){8-11}
& & Method
& \textit{HellaSwag} & \textit{PIQA} & \textit{MMLU} & \others{A{vg.\,$\downarrow$}}
& \textit{HellaSwag} & \textit{PIQA} & \textit{MMLU} & \others{{Avg.\,$\downarrow$}} \\
\midrule

% Calibration-free
\multirow{10}{*}{\rotatebox[origin=c]{90}{\scriptsize\textsc{Calibration-free}}}
& \multirow{4}{*}{\rotatebox[origin=c]{90}{\scriptsize\textsc{3-bit}}}
   & RTN\tnote{$\dagger$}            & 8.44 & 8.60 & 10.97 & \others{9.34} & 22.84 & 17.08 & 10.61 & \others{16.84} \\
&  & Hadamard + RTN\tnote{$\dagger$} & 10.68 & 10.93 & 16.21 & \others{12.60} & 19.83 & 13.17 & 12.81 & \others{15.27} \\
&  & HQQ            & 7.99 & 7.94 & 14.28 & \others{10.07} & 7.23 & 9.30 & 10.98 & \others{9.17} \\
&  & \ours{\textbf{SINQ} \textit{(ours)}} & \ours{\bfseries 5.34} & \ours{\bfseries 7.02} & \ours{\bfseries 10.82} & \ours{\bfseries 7.73} & \ours{\bfseries 5.54} & \ours{\bfseries 7.13} & \ours{\bfseries 10.21} & \ours{\bfseries 7.63} \\
\cmidrule(lr){2-11}
& \multirow{6}{*}{\rotatebox[origin=c]{90}{\scriptsize\textsc{4-bit}}}
  & RTN\tnote{$\dagger$}             & 2.92 & 4.57 & 4.89 & \others{4.13} & 4.18 & 6.31 & 5.28 & \others{5.26} \\
& & BnB (FP4)       & 4.21 & 5.71 & 6.72 & \others{5.55} & 12.32 & 9.14 & 6.25 & \others{9.24} \\
& & BnB (NF4)       & 2.66 & {\bfseries 3.10} & 4.70 & \others{3.49} & 3.73 & 3.48 & 4.76 & \others{3.99} \\
&  & Hadamard + RTN\tnote{$\dagger$}  & 3.63 & 5.55 & 4.88 & \others{4.69} & 4.01 & 6.02 & 5.32 & \others{5.12} \\
&  & HQQ             & 2.81 & 4.35 & 5.17 & \others{4.11} & 5.83 & 5.18 & 4.98 & \others{5.33} \\
&  & \ours{\textbf{SINQ} \textit{(ours)}} & \ours{\bfseries 2.36} & \ours{3.37} & \ours{\bfseries 4.65} & \ours{\bfseries 3.46} & \ours{\bfseries 2.52} & \ours{\bfseries 3.59} & \ours{\bfseries 4.69} & \ours{\bfseries 3.60} \\
\midrule
% Calibrated
\multirow{7}{*}{\rotatebox[origin=c]{90}{\scriptsize\textsc{Calibrated}}}
& \multirow{3}{*}{\rotatebox[origin=c]{90}{\scriptsize\textsc{3-bit}}}
   & GPTQ             & 5.18 & 7.83 & 11.17 & \others{8.06} & 6.33 & 8.76 & 10.25 & \others{8.45} \\
&  & Hadamard$^\dagger$   + GPTQ  & 5.14 & 7.56 & 11.15 & \others{7.95} & 5.52 & 8.71 & {\bfseries 10.08} & \others{8.10} \\
%&  & SpQR  & -- & -- & -- & \others{--} & -- & -- & -- & \others{--} \\
&  & \ours{\textbf{A-SINQ} \textit{(ours)}} & \ours{\bfseries 5.13} & \ours{\bfseries 7.18} & \ours{\bfseries 10.36} & \ours{\bfseries 7.56} & \ours{\bfseries 5.23} & \ours{\bfseries 7.62} & \ours{10.15} & \ours{\bfseries 7.67} \\
\cmidrule(lr){2-11}
& \multirow{4}{*}{\rotatebox[origin=c]{90}{\scriptsize\textsc{4-bit}}}
   & GPTQ              & 2.24 & 4.13 & 4.56 & \others{3.64} & 2.78 & {\bfseries 3.48} & 4.80 & \others{3.69} \\
&  & Hadamard$^\dagger$   + GPTQ   & 2.22 & 3.54 & 4.53 & \others{3.43} & 2.70 & 3.54 & 4.79 & \others{3.68} \\
%&  & SpQR  & -- & -- & -- & \others{--} & -- & -- & -- & \others{--} \\
&  & AWQ               & 2.23 & 3.26 & {\bfseries 4.10} & \others{3.20} & 2.59 & 4.13 & 4.44 & \others{3.72} \\
&  & \ours{\textbf{A-SINQ} \textit{(ours)}} & \ours{\bfseries 2.20} & \ours{\bfseries 3.11} & \ours{4.23} & \ours{\bfseries 3.18} & \ours{\bfseries 2.57} & \ours{3.86} & \ours{\bfseries 4.38} & \ours{\bfseries 3.60} \\
\bottomrule
\bottomrule
\end{tabular}
\begin{tablenotes}
  \item[$\dagger$] Baseline result obtained by running our own implementations.
\end{tablenotes}
\end{threeparttable}
\end{table*}

% A simple explanation for why uncalibrated SINQ can match or sometimes exceed the quality of calibrated methods is that the input scales ($\vec{t}$ in Eq.~\ref{eq:vector-grouping}) SINQ finds are highly correlated with the sample mean of the absolute input activations, i.e. $\vec{\mu_x}$ in Sec.~\ref{sec:awq}, see Fig.~\ref{fig:correlation}. This $\vec{\mu_x}$ is the prescaling factor AWQ \cite{AWQ} uses to improve quantization accuracy. Therefore, a plausible partial interpretation of SINQ is that it is a calibration-free approximation to activation-aware quantization. In contrast to AWQ, SINQ does not require calibration data or parameter fitting.

% \begin{figure}
%     \centering
%     \includegraphics[width=0.6\linewidth]{figures/solution_alignment_evidence.pdf}
%     \caption{Correlation between the average input magnitude and the input scales found by uncalibrated SINQ over different layers of Qwen3-1.7B (compared to a shuffled SINQ scale vector as a baseline). The average is at 0.42.}
%     \label{fig:correlation}
% \end{figure}

% $\text{Corr}(\vec{\mu_x}, \vec{t}_\text{SINQ})$

\subsubsection{Results on Large Models}
%{\color{red} Here we want results on DS-V2, DS-V3 (only ours and HQQ), Qwen3-235B-A22B, ? using ours, HQQ, AWQ(?), BNB.}
We further evaluate our method on two large models, Qwen3-235B-A22B by \cite{qwen3} and DeepSeek-V2.5-236B \cite{DS-V2}, see Tab.~\ref{tab:deepseek-qwen-ptq} in the appendix. Notably, these are both MoE models, and the latter uses Multi-head Latent Attention (MLA). This underlines the robustness of SINQ to different architectures. 

% \input{tables/split_table_test}
% \paragraph{Reasoning} Also in Tab.~\ref{tab:deepseek-qwen-ptq} we report AIME-2025 (\cite{aime}) accuracy for Qwen3-14B. Here we observe a large accuracy drop for prior methods, while matching BF16 performance with SINQ.

\subsection{Uncalibrated Non-Uniform Quantization}
\begin{table*}[t]
\centering
\begin{threeparttable}
\small
\setlength{\tabcolsep}{3.6pt}
\caption{Weight-only uncalibrated PTQ on Qwen3 models with 4-bit non-uniform quantization, reporting perplexity and actual memory usage (GB). Lower is better for all metrics. 
The best \emph{non-uniform} result for a given setting is marked in \textbf{bold}, the results where SINQ with uniform quantization outperforms the non-uniform baselines are marked \textcolor{red}{red}.}
\label{tab:qwen-ptq-nonuniform}
\begin{tabular}{l@{} l l*{9}{r}}
\toprule
& & & \multicolumn{3}{c}{\textbf{Qwen3-1.7B}} & \multicolumn{3}{c}{\textbf{Qwen3-14B}} & \multicolumn{3}{c}{\textbf{Qwen3-32B}} \\
\cmidrule(lr){4-6} \cmidrule(lr){7-9} \cmidrule(lr){10-12}
& & Method 
& Mem. & \others{\textit{Wiki2}\,$\downarrow$} & \others{\textit{C4}\,$\downarrow$}
& Mem. & \others{\textit{Wiki2}\,$\downarrow$} & \others{\textit{C4}\,$\downarrow$}
& Mem. & \others{\textit{Wiki2}\,$\downarrow$} & \others{\textit{C4}\,$\downarrow$} \\
\midrule

% Baseline (FP16)
& & Original (BF16) & 3.44 & \others{\textit{16.67}} & \others{\textit{19.21}} & 29.54 & \others{\textit{8.64}} & \others{\textit{12.01}} & 65.52 & \others{\textit{7.60}} & \others{\textit{10.77}} \\
\midrule

& \multirow{5}{*}{\rotatebox[origin=c]{90}{\scriptsize\textsc{4-bit}}}
 & BnB (FP4)             & 1.42 & \others{24.05} & \others{23.44} & 10.59 & \others{8.88} & \others{12.54} & 20.67 & \others{11.93} & \others{16.90} \\
& & BnB (NF4)             & 1.42 & \others{18.00} & \others{20.43} & 10.59 & \others{8.89} & \others{12.27} & 20.67 & \others{7.94} & \others{11.21} \\
& & HIGGS (non-uniform)             & 1.51 & \others{23.98} & \others{25.27} & 10.28 & \others{9.13} & \others{12.56} & 19.88 & \others{8.02} & \others{11.24} \\
& & \ours{\textbf{SINQ} (NF4) \textit{(ours)}} & \ours{1.42} & \ours{\textbf{16.94}} & \ours{\textbf{19.83}} & \ours{10.56} & \ours{\textbf{8.72}} & \ours{\textbf{12.13}} & \ours{20.73} & \ours{\textbf{7.83}} & \ours{\textbf{10.97}} \\
& & \ours{\textbf{SINQ} \textit{(ours, uniform)}} &  \ours{1.42} & {\color{red}\ours{17.14}}& {\color{red}\ours{19.83}} & \ours{10.56} & {\color{red}\ours{8.76}} & {\color{red}\ours{12.21}} & \ours{20.73} & {\color{red}\ours{7.74}} & {\color{red}\ours{10.96}} \\
\bottomrule
\bottomrule
\end{tabular}
\end{threeparttable}
\end{table*}

\begin{table*}[t]
\centering

\begin{threeparttable}
\small
\setlength{\tabcolsep}{3.6pt}
\caption{Weight-only PTQ on Qwen3 models with 3-bit and 4-bit quantization, reporting perplexity and actual memory usage (GB). Lower is better for all metrics. 
The best result for a given setting is marked in \textbf{bold}, the \emph{calibration-free} results that outperform all calibrated baselines at equal bits (other than our own) are marked \textcolor{red}{red}.
% In bold is the best result for a given setting; in red are calibration-free results that outperform all calibrated baselines at equal bits (other than our own). 
}
\label{tab:qwen-ptq-calibrated}
\begin{tabular}{l@{} l l*{9}{c}}
\toprule
& & & \multicolumn{3}{c}{\textbf{Qwen3-1.7B}} & \multicolumn{3}{c}{\textbf{Qwen3-14B}} & \multicolumn{3}{c}{\textbf{Qwen3-32B}} \\
\cmidrule(lr){4-6} \cmidrule(lr){7-9} \cmidrule(lr){10-12}
& & Method 
& Mem. & \others{\textit{Wiki2}\,$\downarrow$} & \others{\textit{C4}\,$\downarrow$}
& Mem. & \others{\textit{Wiki2}\,$\downarrow$} & \others{\textit{C4}\,$\downarrow$}
& Mem. & \others{\textit{Wiki2}\,$\downarrow$} & \others{\textit{C4}\,$\downarrow$} \\
\midrule

% Baseline (FP16)
& & Original (BF16) & 3.44 & \others{\textit{16.67}} & \others{\textit{19.21}} & 29.54 & \others{\textit{8.64}} & \others{\textit{12.01}} & 65.52 & \others{\textit{7.60}} & \others{\textit{10.77}} \\
\midrule

% Calibrated
% \multirow{7}{*}{\rotatebox[origin=c]{90}{\scriptsize\textsc{Calibrated}}}
& \multirow{4}{*}{\rotatebox[origin=c]{90}{\scriptsize\textsc{3-bit}}}
   & GPTQ               & 1.26 & \others{32.21} & \others{31.05} & 9.28 & \others{9.54} &  \others{13.03} &  17.70 & \others{9.03} & \others{12.38} \\
& & Hadamard$^\dagger$   + GPTQ      & 1.26 & \others{24.70} & \others{25.37} & 9.28 & \others{9.61} &  \others{12.92} &  17.70 & \others{8.51} & \others{11.63} \\
%& & SpQR                 &  --  & \others{--   } & \others{--   } &   -- & \others{--  } &  \others{--}    &  --    & \others{\textbf{--}}   & \others{--   } \\
%& & AWQ$^*$           & 1.26 & 21.87 & 23.17 & 8.90 & 9.41 & 12.76 & 16.68 & 8.36 & -- \\
& & \ours{\textbf{A-SINQ} \textit{(ours)}} & \ours{1.26} & \ours{\textbf{22.30}} & \ours{\textbf{24.00}} & \ours{8.90} & \ours{\textbf{9.31}} & \ours{\textbf{12.71}} & \ours{16.68} & \ours{\textbf{8.45}} & \ours{\textbf{11.54}} \\
& & \ours{\textbf{SINQ} \textit{(ours, calibration-free)}} & \ours{1.28} &\color{red} \ours{22.39} & \color{red}\ours{24.88} & \ours{9.25} & \color{red}\ours{9.33} &\color{red} \ours{12.90} & \ours{17.61} & \ours{8.79} & \ours{11.83} \\
\cmidrule(lr){2-12}
\cmidrule(lr){2-12}
& \multirow{5}{*}{\rotatebox[origin=c]{90}{\scriptsize\textsc{4-bit}}}
   & GPTQ               & 1.38 & \others{19.70} & \others{21.51} & 10.24 & \others{8.81} & \others{12.22} & 19.99 & \others{7.80} & \others{10.99} \\
& & Hadamard$^\dagger$   + GPTQ        & 1.38 & \others{18.12} & \others{20.38} & 10.24 & \others{8.81} & \others{12.19} & 19.99 & \others{7.78} & \others{10.95} \\
%& & SpQR                   &  --  & \others{18.85} & \others{21.37} &   --  & \others{--  } &  \others{--}    &  --  & \others{\textbf{--}}   & \others{--   } \\
& & AWQ               & 1.38 & \others{16.90} & \others{19.95} & 10.25 & \others{8.78} & \others{12.24} & 20.00 & \others{7.79} & \others{10.96} \\
& & \ours{\textbf{A-SINQ} \textit{(ours)}} & \ours{1.38} & \ours{\textbf{16.67}} & \ours{\textbf{19.73}} & \ours{10.21} & \ours{\textbf{8.71}} & \ours{\textbf{12.13}} & \ours{19.83} & \ours{7.78} & \ours{\textbf{10.93}} \\
    & & \ours{\textbf{SINQ} \textit{(ours, calibration-free)}} & \ours{1.42} & \ours{17.14} & \ours{\color{red} 19.83} & \ours{10.58} & \ours{8.76} & \ours{12.21} & \ours{20.73} & \ours{ \textbf{\color{red}7.74}} & \ours{10.96} \\

    \bottomrule
\bottomrule
\end{tabular}
\begin{tablenotes}
  \item[$\dagger$] Baseline result obtained by running our own implementations.
\end{tablenotes}
\end{threeparttable}
\end{table*}
SINQ is compatible with non-uniform quantization levels, for example, NF4 as defined by \cite{NF4}. In Tab.~\ref{tab:qwen-ptq-nonuniform} we compare to various non-uniform 4-bit quantization methods. We simply replace the RoundToNearest function in Alg.\ref{alg:sinq} with the NF4 quantizer. Also, here the SINQ method improves over the NF4 baseline.  We note that for the 32B model, SINQ with INT4 slightly outperforms SINQ with NF4.

\subsection{Calibrated Uniform Quantization}
To demonstrate compatibility with calibration approaches, in Tab.~\ref{tab:qwen-ptq-calibrated} we consider the combination of SINQ and AWQ (see Sec.~\ref{sec:asinq} for the methodology). For a better match to the original AWQ implementation, we quantize our $\vec{s}, \vec{z}$ to 8 bits in these calibrated experiments.  In several cases, even our uncalibrated method outperforms the calibrated baselines, but the addition of AWQ calibration brings further improvements. 

While our weight structure-based estimation of input scales is effective, calibration data is still useful and provides additional information. %The improvements are especially pronounced for more challenging quantization settings. %As the memory column shows, the additional parameters have a negligible impact in practice. 

\begin{table}[]
\small
    \centering
    \caption{Computational overhead of the additional scale in a naive implementation. We compare the matmul speed of the fast gemlite kernel for W4A16 operation with and without the additional scale as used by SINQ. In practice, this scale can often be absorbed into other operations to reduce overhead further. \textbf{B} indicates batchsize, \textbf{D} the input/output dimension, and \textbf{g}($\cdot$) is the gemlite kernel. }
    \label{tab:gemlite}
    \begin{tabular}{ccccc}
        \toprule
        \textbf{B} & \textbf{D} & \textbf{g($\vec{x}$) [ms]} & \textbf{g($\vec{x}\cdot\vec{t}$) [ms]} & \textbf{Overhead [\%]} \\
        \midrule
        1  & 1024 & 0.0446 & 0.0454 & 1.8\% \\
        1  & 2048 & 0.0448 & 0.0455 & 1.5\% \\
        64 & 1024 & 0.0472 & 0.0476 & 0.8\% \\
        64 & 2048 & 0.0479 & 0.0483 & 0.9\% \\
        \bottomrule
        \bottomrule
    \end{tabular}
\end{table}

\subsection{Inference Time}
\label{sec:inference-time}
The inference time of SINQ-quantized models is very close to, or identical to, that of models quantized with standard methods like HQQ or GPTQ. Specifically, the no-overhead formulation of SINQ (see Sec.~\ref{sec:no-overhead}) achieves identical inference time and we report end-to-end results on \texttt{llama.cpp} in Appendix~\ref{app:gguf}.

For the standard SINQ formulation (with dual-scale overhead), we compare the inference time of a $\text{HQQ}$-quantized linear layer using the $\textbf{gemlite}$ kernel \cite{gemlite} with that of a SINQ-quantized layer. For the latter, we implement the second scale using a PyTorch element-wise multiply before applying the kernel. As shown in Tab.~\ref{tab:gemlite}, this simple approach incurs a negligible $1.8\%$ overhead in the most challenging setting of batch size 1. 
%These $1.8\%$ are expressed in comparison to gemlite W4A16 as a baseline, which is up to 3.8x faster than W16A16 (i.e. SINQ is also much faster than standard W16A16). 
We additionally report end-to-end inference results in \texttt{SGLang}~\cite{sglang} in Tab.~\ref{tab:main-decode-speed}.

%in Appendix~\ref{app:sglange2e} for different models. 

\begin{table}[h]
\centering
\small
\setlength{\tabcolsep}{2pt}
\caption{Baseline W16A16 end-to-end decode throughput on SGLang \cite{sglang} in terms of tokens/s (tps, higher is better) at batch size~1 context length $256$, generation length $512$ as well as speedups over W16A16 for AWQ and SINQ at W4A16.} 
%Although AWQ and SINQ rely on different kernel implementations and are therefore not strictly comparable, both achieve over $2\times$ higher end-to-end decode throughput than the FP16 (W16A16) baseline, demonstrating the practical inference benefits of W4A16 quantization.}

\label{tab:main-decode-speed}
\begin{tabular}{l c c c c}
\toprule
 Method 
& \textbf{Llama2-7B} 
& \textbf{Llama3.1-8B} 
& \textbf{Qwen3-14B} 
& \textbf{Qwen3-32B} \\
\midrule
FP16    & 96$tps$ & 86$tps$ & 48$tps$ & 21$tps$ \\
\midrule
AWQ  & 2.4$\times$ & 2.2$\times$ & 2.4$\times$ & 2.9$\times$ \\
\ours{SINQ } & \ours{2.3$\times$} & \ours{2.1$\times$} & \ours{2.4$\times$} & \ours{2.8$\times$} \\
\bottomrule
\bottomrule
\end{tabular}
\end{table}

%This inference time can be further reduced by writing a custom kernel that includes the second scale. However, given the small overhead of the naive implementation, the required engineering effort is disproportionate. 

\subsection{Quantization Time}
Quantization with SINQ is fast. 
%Even on a single recent GPU, quantizing Qwen3-32B takes less than one minute. 
On identical hardware, SINQ has an average runtime of $1.1\times$ our RTN baseline. This is faster than the already efficient HQQ, at $>2\times$, or calibrated methods like AWQ, at $>30\times$ the RTN baseline.
Further details are given in Tab.~\ref{tab:qwen-time} and Fig.~\ref{fig:quant_time} in the appendix.
%\include{tables/main_quant_time}

% I want these experiments:
% \begin{itemize}
%     \item Wikitext and C4 PPL for many models (Qwen3, Llama2/3/4, DeepSeek V2-Lite/V2.5/V3)  (both calibrated/non calibrated setting)
%     \item Flips count (both calibrated/non calibrated setting)
%     \item Actual memory occupation and quantization time
%     \item (Additional) FP4/NVFP4/MXFP4 results
%     \item (Additional) Results on a Vision Model (e.g., ViT or VLM)
%     \item (Additional) Exp on a reasoning bench (reasoning trace length)  
%     \item (Additional) Mixed-precision results
%     \item (optional) Error heatmap of RTN vs SINQ for a matrix with a  single outlier
%     \item (optional) Convergence analysis and time measurement of sinkhorn scaling
% \end{itemize}
% \input{tables/main_ppl}

% {
% \vspace{2pt}
% \raggedright\footnotesize
% \textbf{Notes.} For \emph{Original (FP16)}, fill absolute values:
% Len = avg. reasoning length (tokens), Acc. = accuracy in \%.
% For all quantized rows, report deltas vs. Original:
% \(\Delta\)Len in tokens (e.g., \(-85\) = 85 fewer tokens);
% \(\Delta\)Acc in percentage points (e.g., \(+1.2\) = +1.2 pp).
% Averages are across AIME'23, AIME'24, and MATH500.
% Memory (Mem.) reflects inference memory footprint.
% \end{table}
% }

% \subsection{Qwen3 Output Quality}

% \subsection{Generalization of Calibration}
% -Comparisons PPL and flips GPTQ/AWQ compared to 

\subsection{Ablation Studies}
\begin{figure}[h]   % [t] -> top of column (ICLR likes top floats)
  \centering
  % 0.23\textwidth leaves ~2 % for separation; tweak for exact fit
  \subfloat[]{\includegraphics[width=0.24\textwidth]{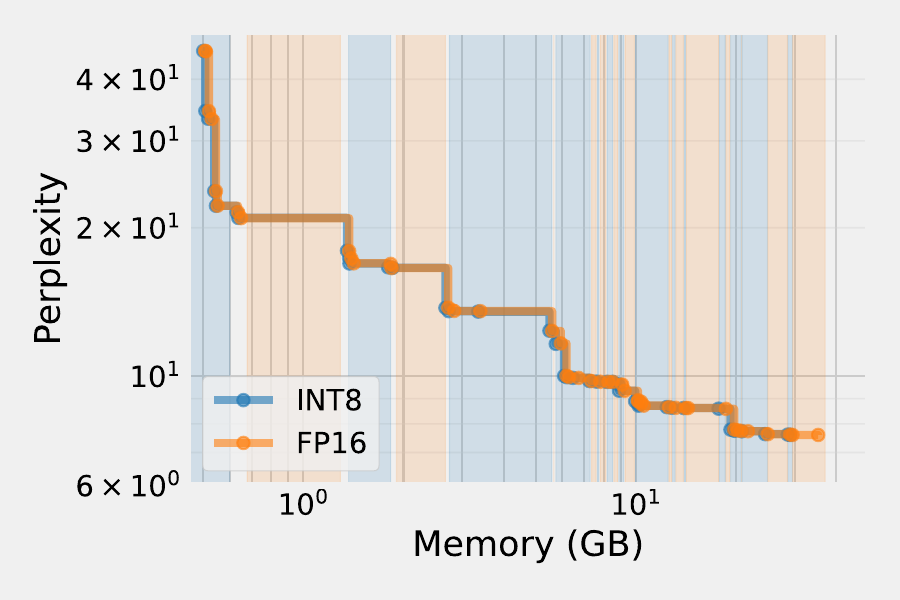}\label{fig:4s_a}}
  %\hfill              % equal spacing between sub-floats
  %\subfloat[]{\includegraphics[width=0.24\textwidth]{figures/ablations/pareto_tiling_dim.pdf}\label{fig:4s_b}}
    \hfill              % equal spacing between sub-floats
  \subfloat[]{\includegraphics[width=0.24\textwidth]{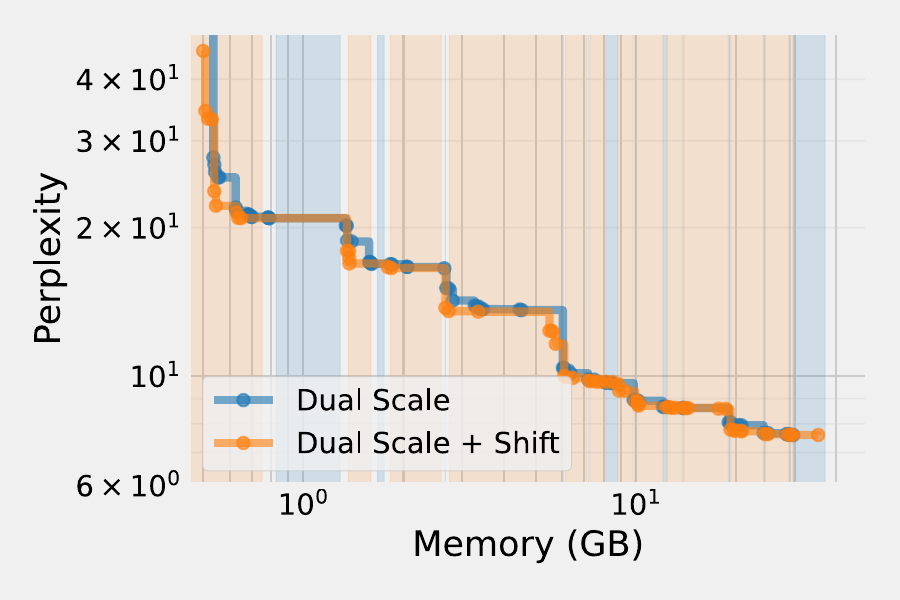}\label{fig:4s_b}}
  \caption{Ablation experiments in the form of memory-perplexity Pareto-fronts across the Qwen3 family. (a) Auxiliary variable precision 
  %(b) Tiling dimension 
  (b) Using or not using shifts. }
  \label{fig:ablation}
\end{figure}

We compare several variants of our method, 1) with and without shifts, 
%2) 1D and 2D tiling\footnote{With a dual-scaling method (in contrast to prior methods) 2D tiling is a non-trivial alternative to 1D tiling. We do not go into detail, because we have not found it to be useful.}, 
2) quantized (int8) and half precision (fp16) auxiliary variables. In Fig.~\ref{fig:ablation}, we see that in general, both  precisions work well and both settings have their sections of the Pareto front. The use of shifts does improve the Pareto front appreciably. Based on these results, we use shifts and  
%choose a 1D tiling with shifts as a good default setting and 
quantize the auxiliaries to match the methods we are comparing against.

% {\color{red} Here we put lots of pareto plots comparing our own methods against each other: With zeros and without, 1D vs. 2D tiling, quantAux vs no quantAux, others?}

% \input{tables/app_llamappl}

\section{Related Work}
\label{sec:related}
\subsection{Uncalibrated, Uniform Integer Quantization}
Most closely related to our approach are works focusing on quantization to uniform integer values without the use of a calibration set. Beyond the trivial (but effective) round-to-nearest (RTN) method with scales and shifts chosen to cover the full range of the input weights, there have been two major innovations in this domain. Firstly, half-quadratic quantization (HQQ, \cite{HQQ}) proposes optimizing the values of the shifts found by RTN, so that a $p$-norm (usually $p=0.7$) error between the original and the quantized matrix becomes minimal. Secondly, applying a Hadamard transform to all weights in a network has been observed to normalize the weight distributions \cite{hadamard}, which often eases quantization. The Hadamard approach has a high-level similarity to our approach, in that we also transform the weight matrices to find an easier-to-quantize format.

% HQQ, RTN, Hadamard, 

% \paragraph{Quantization Space transformations}
% Hadamard, smooth quant (activations), awq, 

\subsection{Non-Uniform Quantization}
After training, neural network weights are usually not uniformly distributed. Therefore, quantization incurs lower errors when the quantization levels are also non-uniform, to match the distribution of the trained weights. \cite{NF4} proposes quantiles of the normal distribution as a preferable set of quantization levels resulting in the normal-float-4 (NF4) format (in the 4-bit case). The variance between optimal levels across different layers in a network is reduced when the weights of the network have been Hadamard transformed. This is used in HIGGS by \cite{HIGGS} together with non-uniform quantization: Non-uniform quantization levels can be synergistic with weight matrix transformations. SINQ is orthogonal to the uniformity of the quantization levels; we show that it is compatible with non-uniform quantization in NF4-based experiments. %The evident downsides of non-uniform approaches are that a lookup needs to be performed and that computation cannot operate at low precision, even when combined with activation quantization. 

% NF4, (HIGGS?), ?
% \begin{itemize}
%     \item need look-up (slow)
%     \item cannot use int4/int8 matmuls when combined with activation quant
% \end{itemize}

\subsection{Calibration}
If quantization time and potential overfitting can be tolerated, using some data to calibrate the quantized value assignments can be a practical approach. A highly influential work is GPTQ \cite{GPTQ} that considers the Hessian for a given layer to find weight pairs that can compensate for each other, if their quantization errors have opposite signs. A second approach, as seen in AWQ \cite{AWQ}, is to minimize the prediction error of each linear layer (separately) under quantization (for more details see Sec.~\ref{sec:awq}). This per-layer prediction error minimization has been further developed by \cite{omniquant} and \cite{affinequant}. Similar to AWQ, CrossQuant \cite{liu2024crossquant} finds an input axis scale for the weight matrix with a calibration process. \cite{any4} combine non-uniform quantization with calibration to learn optimal non-uniform quantization levels. We demonstrate the compatibility of SINQ with calibration in AWQ-based experiments.

%.awq, .GPTQ, quip, .affinequant, .omniquant, spinquant

\subsection{Weight Space Transformations}
The concept of weight space transformation, such as applying the Hadamard transform, a random rotation, or scaling with a diagonal matrix, can be further improved by combining it with calibration and/or non-uniform quantization.  HIGGS \cite{HIGGS} applies Hadamard transforms and matches non-uniform quantization levels to the typically resulting distribution. QuaRot \cite{quarot}, SpinQuant \cite{spinquant}, and FlatQuant \cite{flatquant} combine various calibration methods with rotations (including the Hadamard transform). Duquant \cite{duquant} combines learned rotations with permutations for further flexibility. In Kurtail, \cite{kurtail} optimize rotations on a kurtosis proxy target. Several of these methods specifically target joint activation and weight quantization. The key differences to our method are that we use the dual-scaling and minimize the matrix imbalance, allowing the method to be uniform, calibration-free and, compared to rotated models, architecture agnostic (similar to HQQ \cite{HQQ} and BnB \cite{NF4}) in the sense that each linear layer can be treated independently (which is helpful for generalization to new architectures).

% awq, quarot, spinquant, flatquant, HIGGS, Duquant

% \subsection{Others}

% Codebook-based (AQLM), 
% Non-uniform precision allocation (AdpQ)
% Vector based (VPTQ)
% trellis coded quantization (QTIP)
% low-rank residuals (RESQ)

\section{Conclusion}
\label{sec:conclusion}
We have shown that the column-wise standard deviations of LLM weight matrices are predictive of input activations. To enable the use of this pseudo-activation-awareness for uncalibrated weight quantization, we have proposed a modified Sinkhorn iteration procedure that normalizes both row and column standard deviations, which successfully balances a pseudo-activation-aware column scaling, against row-wise kurtosis impact. We show in numerous experiments that this method is fast and outperforms state-of-the-art methods for uniform quantization without calibration, and can be combined with widely used calibrated and/or non-uniform methods. Finally, we demonstrate that the computational overhead of an additional scale is negligible at the individual-layer level and in end-to-end measurements.

\appendix
\section{Appendix}
\subsection{Impact Statement}
 This paper presents work whose goal is to advance the field of machine learning. There are many potential societal consequences of our work, none of which we feel must be specifically highlighted here.

\subsection{Reproducibility}
The code used to derive our LLM quantization results is given in the supplementary. This includes a full implementation of our method. For our key results, the perplexity evaluations, we use open-source code by \cite{qwenquant} to ensure reproducible detail settings (e.g., context length). Our code, as well as the external code we base ours on, is permissively licensed to facilitate follow-up research. For our experiments, we use permissively licensed open-weight models to promote reproducibility further. %Our Code is given to reproduce the key results in this paper.

\subsection{Weight Standard Deviation Plots for Further Models}
\begin{figure}
    \centering
    \begin{subfigure}{0.45\linewidth}
        \centering
        \includegraphics[width=\linewidth]{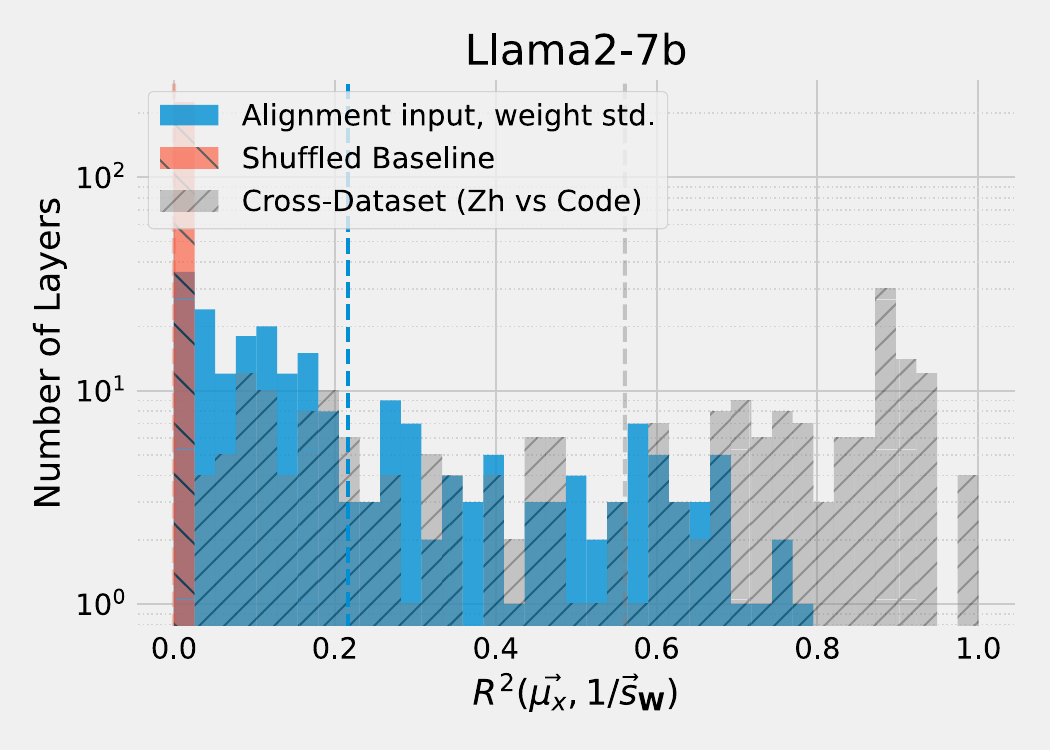}
        \caption{Llama2-7b}
        \label{fig:corr_llama}
    \end{subfigure}
    \begin{subfigure}{0.45\linewidth}
        \centering
        \includegraphics[width=\linewidth]{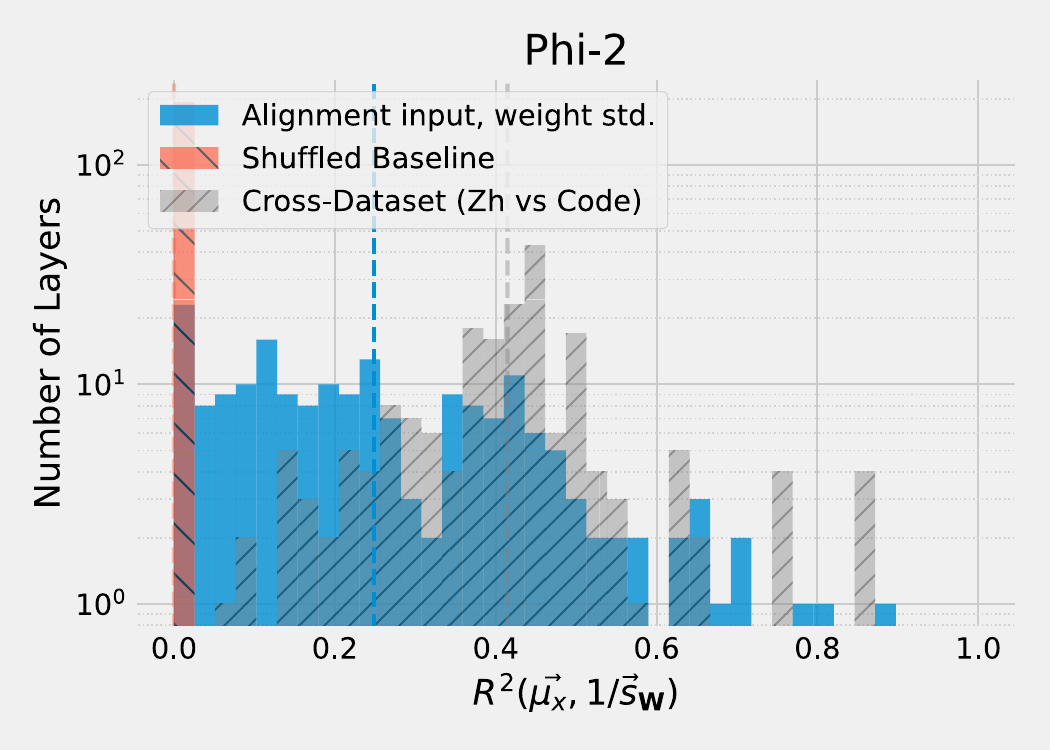}
        \caption{Phi-2}
        \label{fig:corr_phi}
    \end{subfigure}
    \begin{subfigure}{0.45\linewidth}
        \centering
        \includegraphics[width=\linewidth]{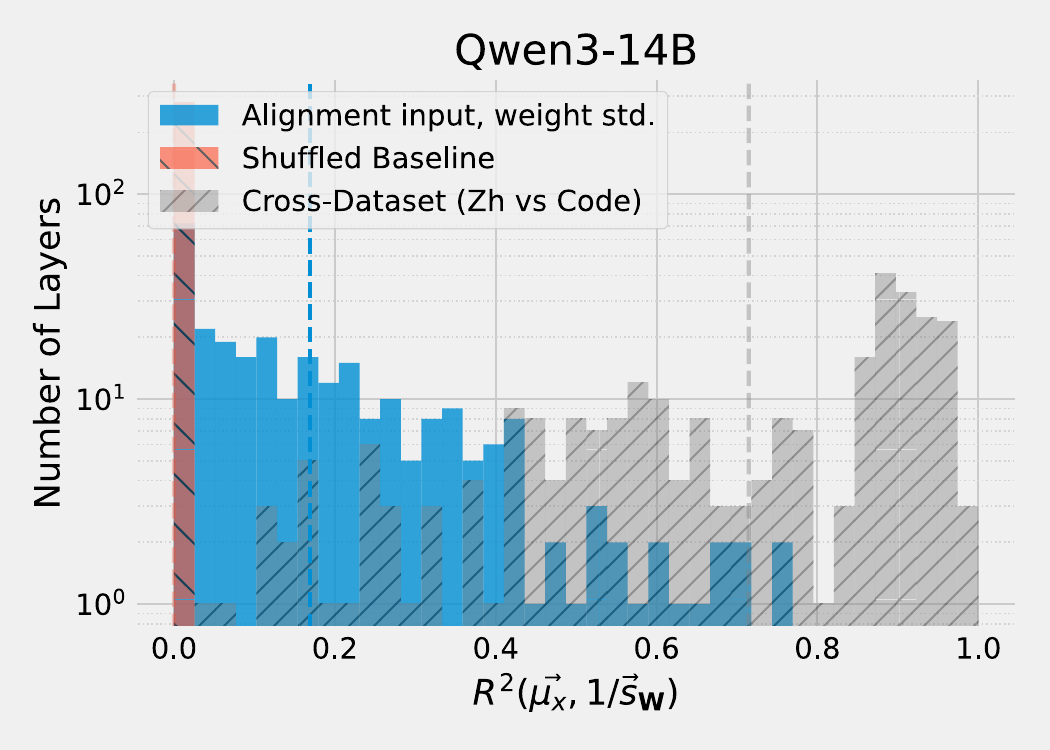}
        \caption{Qwen3-14B}
        \label{fig:corr_qwen14}
    \end{subfigure}
    \begin{subfigure}{0.45\linewidth}
        \centering
        \includegraphics[width=\linewidth]{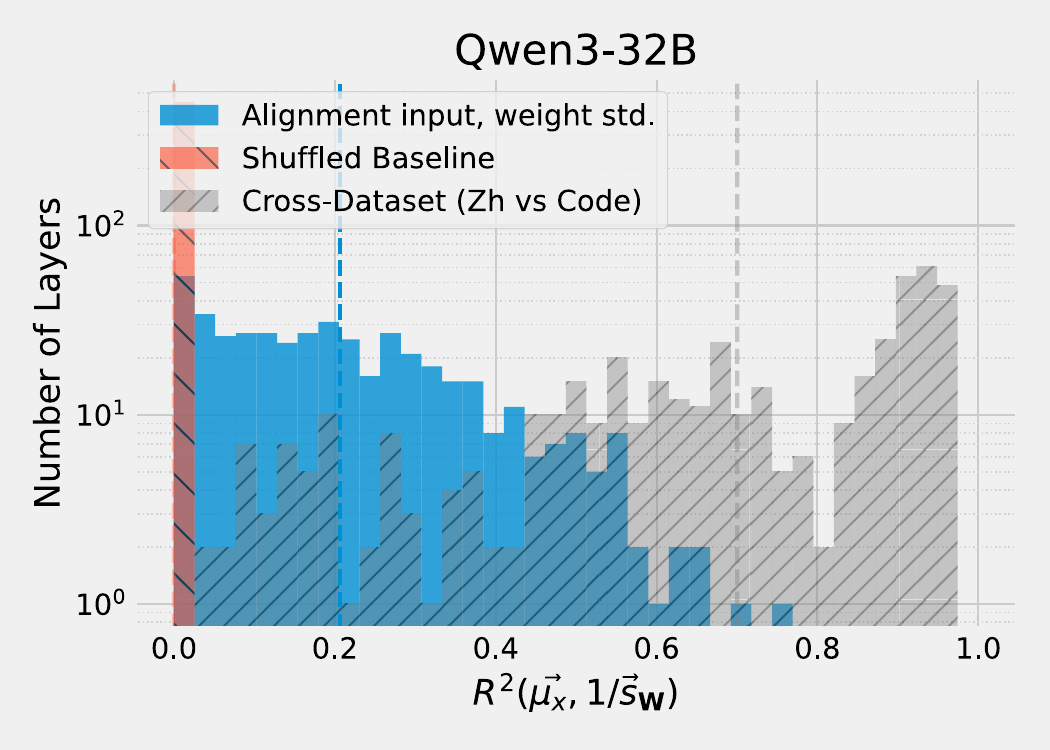}
        \caption{Qwen3-32B}
        \label{fig:corr_qwen32}
    \end{subfigure}
    \caption{$R^2$-Values between activations and weight standard deviations (column-wise) in different models. }
    \label{fig:corr_combined}
\end{figure}

In Fig.~\ref{fig:corr_combined} we see that also in LLama2, Phi-2 and further Qwen3 models, the column-wise weight standard deviation and the input magnitude show a high $R^2$-value.

\subsection{AWQ vs. ASINQ Row-Wise Kurtosis}
\begin{figure}
    \centering
    \includegraphics[width=0.9\linewidth]{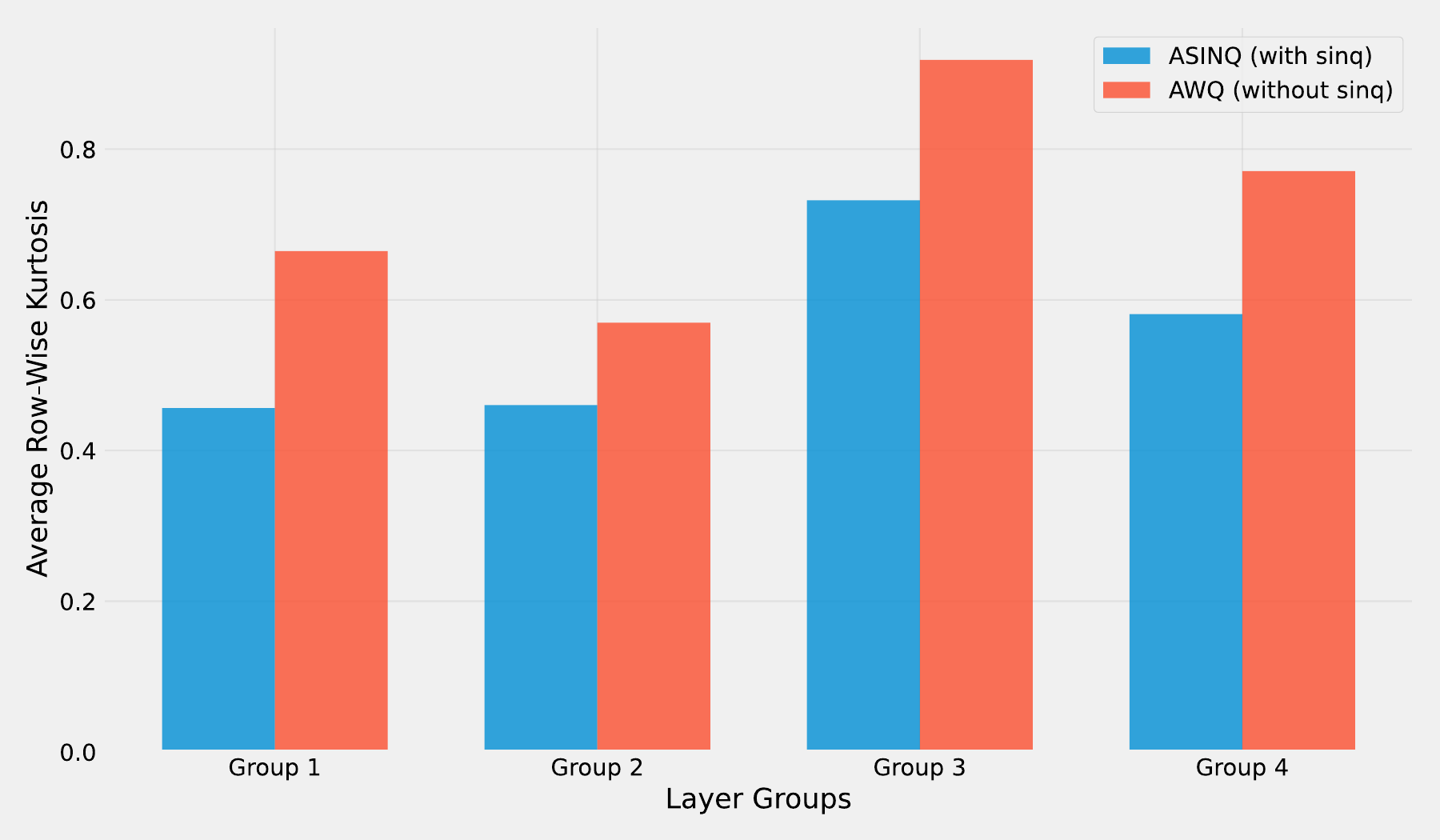}
    \caption{Average row-wise kurtosis for AWQ and ASINQ across different layer groups in Qwen3-1.7B. Using SINQ prevents kurtosis increase in AWQ scaling.}
    \label{fig:awq-asinq-kurt}
\end{figure}

In Fig.~\ref{fig:awq-asinq-kurt} we show the average row-wise kurtosis for AWQ and ASINQ on Qwen3-1.7B. We see that using SINQ before AWQ reduces the row-wise kurtosis, which explains the better quantized performance. The average kurtosis reduction is $1.32\times$.

\subsection{Results on Reasoning}
\label{sec:app:reason}
\begin{table*}[]
\centering
\small
\setlength{\tabcolsep}{6.5pt} % adjusted spacing
\caption{Reasoning performance on Qwen3-14B with 4-bit weight-only PTQ.
}
\label{tab:qwen-reasoning-ptq}
\begin{tabular}{l l l *{6}{c}}
\toprule
& & & \multicolumn{6}{c}{\textbf{Qwen3-14B}} \\
\cmidrule(lr){4-9}
& & Method
& \multicolumn{2}{c}{\textit{AIME 2024}}
& \multicolumn{2}{c}{\textit{AIME 2025}}
& \multicolumn{2}{c}{\others{Avg.}} \\
\cmidrule(lr){4-5} \cmidrule(lr){6-7} \cmidrule(lr){8-9}
& & 
& Tok. & Acc. (\%)\,$\uparrow$
& Tok. & Acc. (\%)\,$\uparrow$
& \others{$\Delta$ Tok.} & \others{Acc. (\%)\,$\uparrow$}\\
\midrule

% Baseline (FP16, absolute values)
& & Original (FP16) 
& \textit{\num{11464}} & \textit{76.70}
& \textit{\num{12636}} & \textit{63.30}
& \others{\textit{0}} & \others{\textit{70.00}} \\
\midrule

% Calibration-free
\multirow{6}{*}{\rotatebox[origin=c]{90}{\scriptsize\textsc{Calibration-free}}}
& \multirow{6}{*}{\rotatebox[origin=c]{90}{\scriptsize\textsc{4-bit}}}
  & RTN                 & \num{10973} & 66.70 & \num{12642} & 50.00 & \others{-242} & \others{58.35} \\
& & BnB (FP4)           & \num{11500} & 60.00 & \num{12455} & 53.30 & \others{-72} & \others{56.65} \\
& & BnB (NF4)           & \num{12132} & 70.00 & \num{12899} & 56.70 & \others{+930} & \others{63.35} \\
& & Hadamard + RTN      & \num{11210} & 70.00 & \num{12989} & 53.30 & \others{+99} & \others{61.65} \\
& & HQQ                 & \num{11862} & 70.00 & \num{12991} & 56.70 & \others{+367} & \others{63.35} \\
& & \ours{\textbf{SINQ}}  & \ours{\num{11660}} & \ours{\textbf{73.30}} & \ours{\num{12305}} & \ours{\textbf{63.30}} & \ours{-67} & \ours{\textbf{68.30}} \\

% % Calibrated
% \multirow{4}{*}{\rotatebox[origin=c]{90}{\scriptsize\textsc{Calibrated}}}
% & & GPTQ                   & --- & --- & --- & --- & \others{---} & \others{---} \\
% & & Hadamard + GPTQ        & --- & --- & --- & --- & \others{---} & \others{---} \\
% & & AWQ                    & --- & --- & --- & --- & \others{---} & \others{---} \\
% & & \ours{\textbf{A-SINQ}} & \ours{---} & \ours{---} & \ours{---} & \ours{---} & \ours{---} & \ours{---} \\
\bottomrule
\bottomrule
\end{tabular}
\end{table*}

In Tab.~\cref{tab:qwen-reasoning-ptq} we show results on reasoning benchmarks \cite{aime}. Here, we include the length of reasoning traces to ensure that lengthened reasoning does not negate some of the upside of quantization. Note that these are preliminary pass@1 results. These preliminary findings seem to suggest that the proposed method sustains robust reasoning capabilities while avoiding an increase in reasoning trace length, which is crucial for preserving the efficiency gains achieved through quantization.

\subsection{No-Overhead variant}
\label{app:nooverhead}
In Tab.~\ref{tab:no-overhead}, we show that the overhead-free formulation of SINQ also produces better quality outputs than comparable prior methods.
\begin{table*}[t!]
\centering
\small
\setlength{\tabcolsep}{3.6pt}
\begin{threeparttable}
\caption{No-Overhead SINQ variant. Weight-only uncalibrated uniform PTQ on Qwen3 models with 4-bit quantization, reporting perplexity and actual memory usage (GB). Lower is better for all metrics.  The best result for a given setting is marked in \textbf{bold}. }
\label{tab:no-overhead}
\begin{tabular}{l l l*{9}{r}}
\toprule
& & & \multicolumn{3}{c}{\textbf{Qwen3-1.7B}} & \multicolumn{3}{c}{\textbf{Qwen3-14B}} & \multicolumn{3}{c}{\textbf{Qwen3-32B}} \\
\cmidrule(lr){4-6} \cmidrule(lr){7-9} \cmidrule(lr){10-12}
& & Method 
& Mem. & \others{\textit{Wiki2}\,$\downarrow$} & \others{\textit{C4}\,$\downarrow$}
& Mem. & \others{\textit{Wiki2}\,$\downarrow$} & \others{\textit{C4}\,$\downarrow$}
& Mem. & \others{\textit{Wiki2}\,$\downarrow$} & \others{\textit{C4}\,$\downarrow$} \\
\midrule

% Baseline (FP16)
& & Original (BF16) & 3.44 & \others{\textit{16.67}} & \others{\textit{19.21}} & 29.54 & \others{\textit{8.64}} & \others{\textit{12.01}} & 65.52 & \others{\textit{7.60}} & \others{\textit{10.77}} \\
\midrule
%& & BnB (FP4)             & 1.42 & \others{24.05} & \others{23.44} & 10.59 & \others{8.88         } & \others{12.54} & 20.67 & \others{11.93} & \others{16.90} \\
%& & BnB(NF4)             & 1.42 & \others{18.00} & \others{20.43} & 10.59 & \others{8.89         } & \others{12.27} & 20.67 & \others{7.94} & \others{11.21} \\
%& & HIGGS (non-uniform)             & 1.51 & \others{23.98} & \others{25.27} & 10.28 & \others{9.13        } & \others{12.56} & 19.88 & \others{8.02} & \others{11.24} \\
& & Hadamard + RTN\tnote{$\dagger$}      & 1.42 & \others{19.10} & \others{20.70} & 10.54 & \others{8.85} & \others{12.35} & 20.78 & \others{8.28} & \others{11.60} \\
& & HQQ             & 1.42 & \others{18.96} & \others{22.10} & 10.54 & \others{8.78} & \others{12.36} & 20.78 & \others{8.62} & \others{12.20} \\
& & \ours{\textbf{SINQ} \textit{(ours)}} & \ours{1.42} & \ours{\textbf{17.14}} & \ours{\textbf{ 19.83}} & \ours{10.56} & \ours{\textbf{8.76}} & \ours{\textbf{12.21}} & \ours{20.73} & \ours{\textbf{7.74}} & \ours{\textbf{10.96}} \\
& & \ours{\textbf{SINQ no overhead} \textit{(ours)}} & \ours{1.42} & \ours{{17.63}} & \ours{{ 19.99}} & \ours{10.56} & \ours{{8.78}} & \ours{{12.32}} & \ours{20.73} & \ours{{7.78}} & \ours{{11.15}} \\
%& & \ours{\textbf{SINQ} (NF4) \textit{(ours)}} & \ours{1.42} & \ours{16.94} & \ours{19.83} & \ours{10.56} & \ours{8.72} & \ours{12.13} & \ours{20.73} & \ours{7.83} & \ours{10.97} \\
\bottomrule
\bottomrule
\end{tabular}
\begin{tablenotes}
  \item[$\dagger$] Baseline result obtained by running our own implementations.
\end{tablenotes}
\end{threeparttable}
\end{table*}

\subsection{Improving GGUF with no-overhead SINQ}
\label{app:gguf}
Table~\ref{tab:llamacpp} demonstrates that no-overhead SINQ reduces perplexity over standard GGUF while preserving inference speedups, evaluated on \texttt{Q4\_0} and \texttt{Q3\_KS}, the prevalent GGUF formats for 4-bit and 3-bit quantization.
\begin{table*}[t]
\centering
\small
\setlength{\tabcolsep}{5.8pt} % adjusted spacing
\caption{Performance of 4-bit and 3-bit no-overhead SINQ combined with GGUF quantization in \texttt{llama.cpp} on GPU, evaluated at batch size $512$ (input length $512$, output length $128$). We report perplexity (Ppl., lower is better) measured on WikiText-2, as well as prefill and decode throughput (tokens/s, higher is better) measured on a 512-token prompt. Speedups are reported relative to the FP16 baseline. No-overhead SINQ+GGUF consistently improves GGUF quantization.
}
\label{tab:llamacpp}
\begin{tabular}{l *{9}{c}}
\toprule
& \multicolumn{3}{c}{\textbf{Qwen3 0.6B}} & \multicolumn{3}{c}{\textbf{Qwen3 1.7B}} & \multicolumn{3}{c}{\textbf{Qwen3 4B}} \\
\cmidrule(lr){2-4}\cmidrule(lr){5-7}\cmidrule(lr){8-10}
\textbf{Method}
& \others{Ppl.\,$\downarrow$} & Prefill $\uparrow$ & Decode $\uparrow$
& \others{Ppl.\,$\downarrow$} & Prefill $\uparrow$ & Decode $\uparrow$
& \others{Ppl.\,$\downarrow$} & Prefill $\uparrow$ & Decode $\uparrow$ \\
\midrule
\textit{Base (FP16)}
& \others{\textit{21.88}} & \textit{22034 $tps$} & \textit{647 $tps$}
& \others{\textit{17.13}} & \textit{12762 $tps$} & \textit{346 $tps$}
& \others{\textit{14.31}} & \textit{5889 $tps$}  & \textit{167 $tps$} \\
\midrule
Base + Q4\_0
& \others{25.25} & $1.93\times$ & $1.31\times$ 
& \others{21.30} & $2.56\times$ & $1.80\times$ 
& \others{14.97} & $2.92\times$ & $2.10\times$ \\
\ours{No-over. SINQ + Q4\_0}
& \ours{\textbf{24.01}} & \ours{$1.93\times$} & \ours{$1.31\times$}
& \ours{\textbf{17.65}} & \ours{$2.56\times$} & \ours{$1.80\times$}
& \ours{\textbf{14.77}} & \ours{$2.92\times$} & \ours{$2.10\times$} \\
\midrule
Base + Q3\_KS
& \others{35.59} & $1.77\times$ & $1.16\times$
& \others{24.02} & $2.19\times$ & $1.62\times$
& \others{19.03} & $2.55\times$ & $1.80\times$ \\
\ours{No-over. SINQ + Q3\_KS}
& \ours{\textbf{30.20}} & \ours{$1.78\times$} & \ours{$1.16\times$}
& \ours{\textbf{19.49}} & \ours{$2.19\times$} & \ours{$1.62\times$}
& \ours{\textbf{18.06}} & \ours{$2.54\times$} & \ours{$1.80\times$} \\
\bottomrule
\bottomrule
\end{tabular}
\end{table*}

\subsection{Timing Results}
\label{sec:app:time}
\begin{figure}[H]
    \centering
    \includegraphics[width=.8\linewidth]{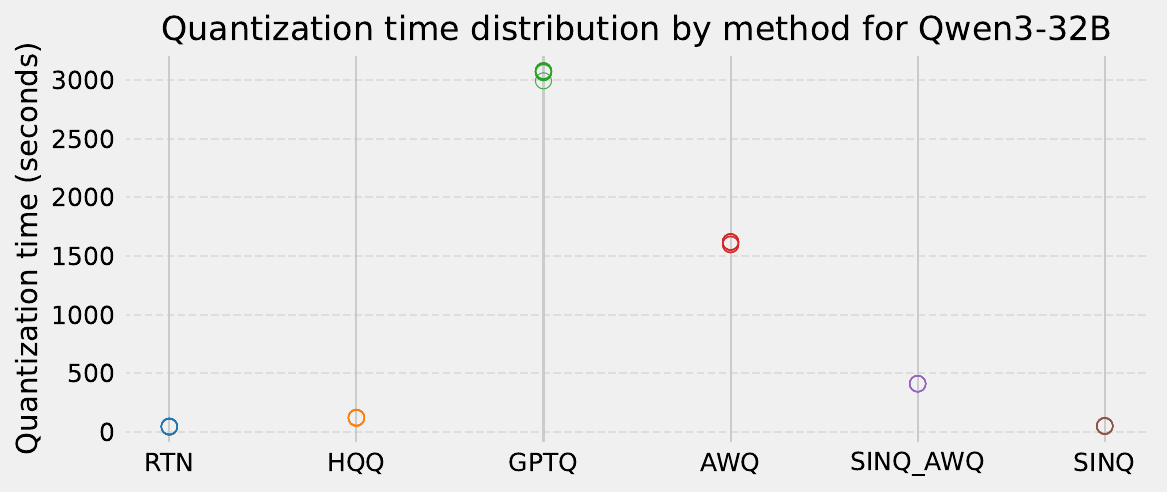}
    \caption{Distribution of quantization times for each method for Qwen3-32B.} %across the quantized Qwen3 models 0.6B, 1.7B, 4B, 8B, 14B, 32B.}
    \label{fig:quant_time}
\end{figure}

In Tab.~\ref{tab:qwen-time} we report quantization time results on a single GPU for various models. Although precise timings may vary with hardware, our method achieves times comparable to the RTN baseline and even surpasses HQQ, which is already regarded as a fast quantization technique. Furthermore, the calibrated version, A-SINQ, is substantially faster than popular state-of-the-art calibrated methods like GPTQ and AWQ. Fig.~\ref{fig:quant_time} shows the distribution of quantization times over 10 runs for various popular quantization methods on Qwen3-32B on GPU.
\begin{table*}[]
\centering
\small
\begin{threeparttable}

\setlength{\tabcolsep}{4pt}
\caption{Average quantization time (seconds) across 10 runs for some Qwen3 models on GPU, comparing different quantization methods. 
The rightmost column reports the relative average slowdown with respect to RTN.}
\label{tab:qwen-time}
\begin{tabular}{l *{5}{r} r}
\toprule
Method 
& \textbf{Qwen3-1.7B} 
& \textbf{Qwen3-4B} 
& \textbf{Qwen3-8B} 
& \textbf{Qwen3-14B} 
& \textbf{Qwen3-32B} 
& \others{Avg. cost} \\
\midrule
\textit{RTN}           & \textit{\SI{2.91}{\second} {\tiny$\pm$0.11}} & \textit{\SI{6.32}{\second} {\tiny$\pm$0.06}} & \textit{\SI{11.35}{\second} {\tiny$\pm$0.31}} & \textit{\SI{20.61}{\second} {\tiny$\pm$0.87}} & \textit{\SI{46.79}{\second} {\tiny$\pm$2.52}} & \others{\textit{1.00$\times$}} \\
\midrule
HQQ & \SI{3.65}{\second} {\tiny$\pm$0.13} & \SI{10.15}{\second} {\tiny$\pm$0.27} & \SI{24.06}{\second} {\tiny$\pm$1.54} & \SI{43.62}{\second} {\tiny$\pm$0.54} & \SI{122.45}{\second} {\tiny$\pm$2.45} & \others{2.32$\times$} \\
GPTQ& \SI{193.33}{\second} {\tiny$\pm$1.68} & \SI{426.89}{\second} {\tiny$\pm$0.75} & \SI{669.06}{\second} {\tiny$\pm$0.84} & \SI{1160.37}{\second} {\tiny$\pm$1.68} & \SI{3064.62}{\second} {\tiny$\pm$24.33} & \others{62.68$\times$} \\
AWQ  & \SI{104.63}{\second} {\tiny$\pm$9.26} & \SI{225.27}{\second} {\tiny$\pm$3.91} & \SI{392.51}{\second} {\tiny$\pm$2.86} & \SI{695.29}{\second} {\tiny$\pm$1.19} & \SI{1613.75}{\second} {\tiny$\pm$9.79} & \others{34.46$\times$} \\
A-SINQ \textit{(ours)} & \SI{23.86}{\second} {\tiny$\pm$0.17} & \SI{49.81}{\second} {\tiny$\pm$0.13} & \SI{92.17}{\second} {\tiny$\pm$0.33} & \SI{173.93}{\second} {\tiny$\pm$0.38} & \SI{411.95}{\second} {\tiny$\pm$0.57} & \others{8.54$\times$} \\
\ours{\textbf{SINQ} \textit{(ours)}} & \ours{\textbf{\SI{3.03}{\second} {\tiny$\pm$0.29}}} & \ours{\textbf{\SI{6.33}{\second} {\tiny$\pm$0.52}}} & \ours{\textbf{\SI{13.23}{\second} {\tiny$\pm$0.64}}} & \ours{\textbf{\SI{21.38}{\second} {\tiny$\pm$2.15}}} & \ours{\textbf{\SI{51.56}{\second} {\tiny$\pm$2.00}}} & \ours{\textbf{1.09$\times$}} \\
\bottomrule
\bottomrule
\end{tabular}
\end{threeparttable}
\end{table*}

\subsection{Results on Llama Models}
\label{sec:llama}
In Tab.~\ref{tab:llama-ptq} we report quantization results on Llama family models. These findings further validate the effectiveness of SINQ also on this type of architecture. 
\begin{table*}[]
\centering
\small
\begin{threeparttable}
\setlength{\tabcolsep}{3.6pt}
\caption{Weight-only PTQ on Llama models with 3-bit and 4-bit quantization, reporting perplexity and actual memory usage (GB). Lower is better for all metrics. In bold is the best result for a given setting. }
\label{tab:llama-ptq}
\begin{tabular}{l l l*{9}{c}}
\toprule
& & & \multicolumn{3}{c}{\textbf{Llama 2-7B}} & \multicolumn{3}{c}{\textbf{Llama 3-8B}} & \multicolumn{3}{c}{\textbf{Llama 3-70B}} \\
\cmidrule(lr){4-6} \cmidrule(lr){7-9} \cmidrule(lr){10-12}
& & Method 
& Mem. & \others{\textit{Wiki2}\,$\downarrow$} & \others{\textit{C4}\,$\downarrow$}
& Mem. & \others{\textit{Wiki2}\,$\downarrow$} & \others{\textit{C4}\,$\downarrow$}
& Mem. & \others{\textit{Wiki2}\,$\downarrow$} & \others{\textit{C4}\,$\downarrow$} \\
\midrule

% Baseline (FP16)
& & Original (BF16) & 14.08 & \others{\textit{5.47}} & \others{\textit{6.90}} & 17.45 & \others{\textit{6.13}} & \others{\textit{9.61}} & 141.11 & \others{\textit{2.86}} & \others{\textit{7.30}} \\
\midrule

% Calibration-free
\multirow{11}{*}{\rotatebox[origin=c]{90}{\scriptsize\textsc{Calibration-free}}}
& \multirow{4}{*}{\rotatebox[origin=c]{90}{\scriptsize\textsc{3-bit}}}
   & RTN                  & 3.54
   & \others{6.40} & \others{8.05} & 5.25 & \others{10.18} & \others{15.27} & 35.93 & \others{5.26} & \others{10.80} \\
&  & Hadamard + RTN       & 3.54 & \others{6.31} & \others{7.89} & 5.25 & \others{9.97} & \others{15.25} & 35.93 & \others{4.99} & \others{10.45} \\
& & HQQ                   & 3.62 & \others{7.05} & \others{9.03} & 5.24 & \others{9.55} & \others{14.68} & 36.16 & \others{85.64} & \others{23.32} \\
& & \ours{\textbf{SINQ} \textit{(ours)}} & \ours{3.54} & \ours{\textbf{6.14}} & \ours{\textbf{7.72}} & \ours{5.35} & \ours{\textbf{8.04}} & \ours{\textbf{12.32}} & \ours{35.93} & \ours{\textbf{4.52}} & \ours{\textbf{8.48}} \\
\cmidrule(lr){2-12}
& \multirow{7}{*}{\rotatebox[origin=c]{90}{\scriptsize\textsc{4-bit}}}
 & RTN              & 4.17 & \others{5.67} & \others{7.14} & 6.06 & \others{6.61} & \others{10.25} & 42.71 & \others{3.56} & \others{10.58} \\
& & BnB (FP4)             & 4.17 & \others{5.76} & \others{7.24} & 6.06 & \others{6.93} & \others{10.75} & 42.71 & \others{3.58} & \others{8.23} \\
& & Hadamard + RTN    & 4.17 & \others{5.65} & \others{7.10} & 6.06 & \others{6.72} & \others{10.23} & 42.71 & \others{3.54} & \others{9.95} \\
& & HQQ             & 4.22 & \others{5.68} & \others{7.13} & 6.06 & \others{6.58} & \others{10.22} & 42.71 & \others{3.26} & \others{8.13} \\
& & \ours{\textbf{SINQ} \textit{(ours)}} & \ours{4.19} & \ours{\textbf{5.60}} & \ours{\textbf{7.04}} & \ours{6.06} & \ours{\textbf{6.53}} & \ours{\textbf{10.14}} & \ours{42.81} & \ours{\textbf{3.17}} & \ours{\textbf{7.51}} \\
\cmidrule(lr){3-12}
& & BnB (NF4)             & 4.17 & \others{5.65} & \others{7.09} & 6.07 & \others{6.56} & \others{10.20} & 42.71 & \others{3.22} & \others{7.68} \\
& & \ours{\textbf{SINQ} (NF4) \textit{(ours)}} & \ours{4.18} & \ours{\textbf{5.58}} & \ours{\textbf{7.03}} & \ours{6.07} & \ours{\textbf{6.51}} & \ours{\textbf{10.09}} & \ours{42.81} & \ours{\textbf{3.16}} & \ours{\textbf{7.50}} \\
%\midrule

% Calibrated
% \multirow{7}{*}{\rotatebox[origin=c]{90}{\scriptsize\textsc{Calibrated}}}
% & \multirow{3}{*}{\rotatebox[origin=c]{90}{\scriptsize\textsc{3-bit}}}
%    & GPTQ                & -- & \others{--} & \others{--} & -- & \others{--} &  \others{--} &  -- & \others{--} & \others{--} \\
% & & Hadamard + GPTQ      & -- & \others{--} & \others{--} & -- & \others{--} &  \others{--} &  -- & \others{--} & \others{--} \\
% & & \ours{\textbf{A-SINQ} \textit{(ours)}} & \ours{3.34} & \ours{6.05} & \ours{7.57} & \ours{5.15} & \ours{7.91} & \ours{12.12} & \ours{--} & \ours{\textbf{--}} & \ours{\textbf{--}} \\
% \cmidrule(lr){2-12}
% & \multirow{4}{*}{\rotatebox[origin=c]{90}{\scriptsize\textsc{4-bit}}}
%    & GPTQ                  & -- & \others{--} & \others{--} & -- & \others{--} & \others{--} & -- & \others{--} & \others{--} \\
% & & Hadamard + GPTQ        & -- & \others{--} & \others{--} & -- & \others{--} & \others{--} & -- & \others{\textbf{--}} & \others{--} \\
% & & AWQ                    & 4.07 & \others{5.58} & \others{7.02} & 5.89 & \others{6.49} & \others{10.07} & -- & \others{--} & \others{--} \\
% & & \ours{\textbf{A-SINQ} \textit{(ours)}} & \ours{3.99} & \ours{5.58} & \ours{7.02} & \ours{5.83} & \ours{6.47} & \ours{10.06} & \ours{} & \ours{\textbf{--}} & \ours{\textbf{--}} \\
\bottomrule
\bottomrule
\end{tabular}
\end{threeparttable}
\end{table*}

\subsection{Results on DeepSeek-V3 and other large models}
\label{sec:app:dsv3}
In Tab.~\ref{tab:dsv3} we compare HQQ to SINQ on WikiText2 perplexity for DeepSeek-V3 \cite{DSV3} while in Tab.~\ref{tab:deepseek-qwen-ptq} we report the results for DeepSeek-V2.5-236B and Qwen3-235B-A22B.
\begin{table}[H]
\centering
\small
\setlength{\tabcolsep}{6pt}
\caption{Weight-only PTQ on \textbf{DeepSeek-V3-685B} with 4-bit quantization.
We report perplexity on WikiText-2 (lower is better). Best per setting in \textbf{bold}.}
\label{tab:dsv3}
\begin{tabular}{l l c}
\toprule
Setting & Method & \textit{Wiki2}\,$\downarrow$ \\
\midrule
Calibration-free (4-bit) & \others{HQQ}         &    \others{5.38}  \\
                         & \ours{SINQ} & \ours{\textbf{5.31}} \\
\bottomrule
\bottomrule
\end{tabular}
\end{table}

% \begin{wrapfigure}{r}{0.5\textwidth}
%   \begin{center}
%     \includegraphics[width=0.48\textwidth]{birds}
%   \end{center}
%   \caption{Birds}
% \end{wrapfigure}
\begin{table*}[t]
\centering
\small
\begin{threeparttable}

\setlength{\tabcolsep}{5pt}
\caption{Weight-only PTQ on \textbf{DeepSeek-V2.5-236B} and \textbf{Qwen3-235B-A22B} MoE models with 3-bit and 4-bit quantization, reporting perplexity and actual memory usage (GB). Lower is better for all metrics. The best result for a given setting is marked in \textbf{bold}. }
\label{tab:deepseek-qwen-ptq}
\begin{tabular}{l l c c c c r r}
\toprule
& & \multicolumn{3}{c}{\textbf{DeepSeek-V2.5-236B}} & \multicolumn{3}{c}{\textbf{Qwen3-235B-A22B}} \\
\cmidrule(lr){3-5} \cmidrule(lr){6-8}
Setting & Method & Mem. & \others{\textit{Wiki2}\,$\downarrow$} & \others{\textit{C4}\,$\downarrow$} 
                       & Mem. & \others{\textit{Wiki2}\,$\downarrow$} & \others{\textit{C4}\,$\downarrow$} \\
\midrule

Baseline & Original (BF16) & 471.56 & \others{5.36} & \others{8.15} 
                           & 470.19 & \others{5.37} & \others{9.30} \\
\midrule

Calibration-free (3-bit) & RTN & 110.90 & \others{5.91} & \others{8.84} 
                                      & 110.98 & \others{10.11} & \others{13.92} \\
                         & HQQ & 110.92 & \others{5.89} & \others{8.76} 
                                      & 114.43 & \others{13.07} & \others{16.38} \\
                        % & \ours{\textbf{SINQ} \textit{(ours, aux. quant.)}} 
                        %       & \ours{104.21} & \ours{\textbf{5.82}} & \ours{} 
                        %       & \ours{--} & \ours{} & \ours{} \\
                         & \ours{\textbf{SINQ} \textit{(ours)}} 
                              & \ours{110.91} & \ours{\textbf{5.82}} & \ours{\textbf{8.74}} 
                              & \ours{110.99} & \ours{\textbf{6.27}} & \ours{\textbf{10.03}} \\
\midrule

Calibration-free (4-bit) & RTN & 134.24 & \others{5.49} & \others{8.27} 
                                      & 134.03 & \others{5.65} & \others{9.49} \\
                         & BnB (FP4)     &  134.52 & \others{5.55} & \others{8.41} 
                                      & 134.10 & \others{6.67} & \others{10.21} \\
                         & BnB (NF4)     &  134.52 & \others{5.49} & \others{8.28} 
                                       & 134.10 & \others{5.60} & \others{9.49} \\
                         & HQQ & 134.25 & \others{5.49} & \others{8.27} 
                                      & 134.03 & \others{5.60} & \others{9.46} \\
                         & \ours{\textbf{SINQ} \textit{(ours)}} 
                              & \ours{134.51} & \ours{\textbf{5.48}} & \ours{\textbf{8.25}} 
                              & \ours{134.06} & \ours{\textbf{5.58}} & \ours{\textbf{9.43}} \\
% \midrule

% Calibrated (3-bit) & \ours{\textbf{A-SINQ} \textit{(ours)}} 
%                               & \ours{--} & \ours{--} & \ours{--} 
%                               & \ours{--} & \ours{--} & \ours{--} \\
% \midrule

% Calibrated (4-bit) & AWQ & -- & \others{--} & \others{--} 
%                                & -- & \others{--} & \others{--} \\
%                    & \ours{\textbf{A-SINQ} \textit{(ours)}} 
%                                & \ours{--} & \ours{\textbf{--}} & \ours{\textbf{--}} 
%                                & \ours{--} & \ours{\textbf{--}} & \ours{\textbf{--}} \\
\bottomrule
\bottomrule
\end{tabular}
\end{threeparttable}

\end{table*}

\subsection{Accuracy Results}
\label{sec:app:acc}
\begin{figure}[t]
    \centering
    \includegraphics[width=.9\linewidth]{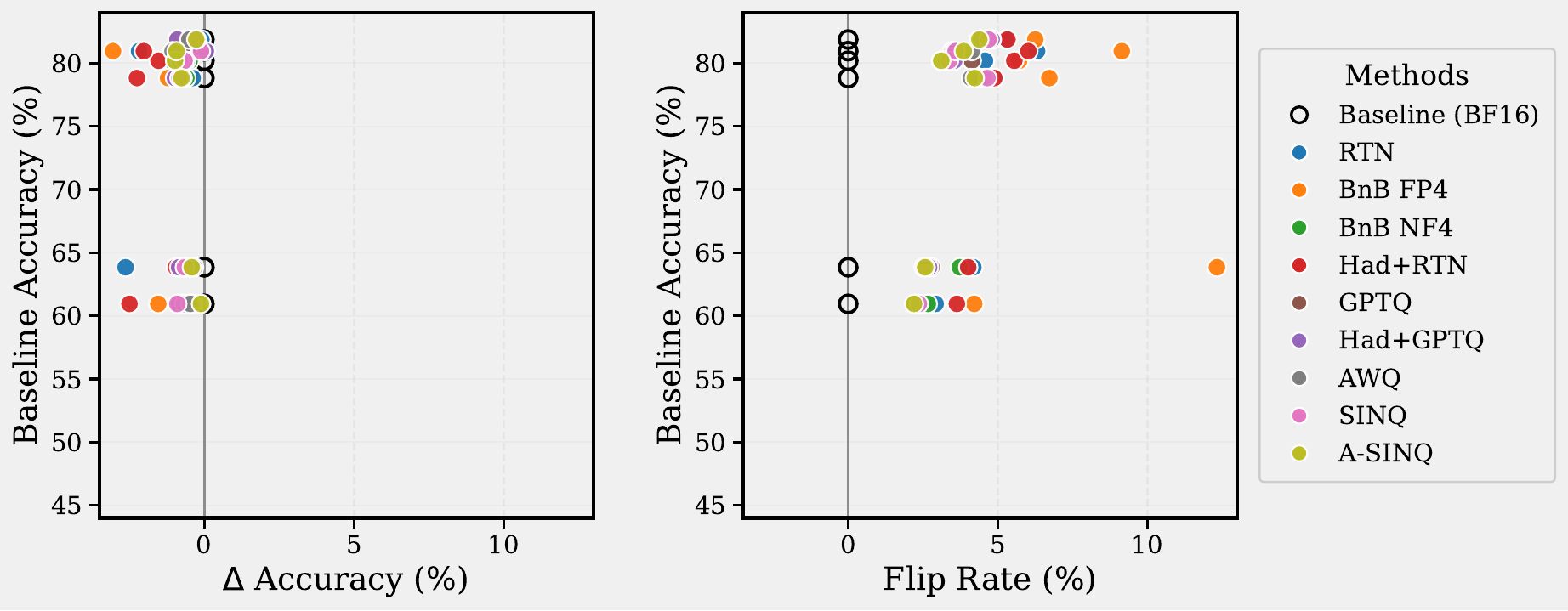}
    \caption{Comparison of baseline accuracy, accuracy changes, and flip rates across different \hbox{4-bit} quantization methods (similar to \cite{notaccuracy}). On QA tasks, flips have been shown to be the more consistent quality metric of LLM quantization.}
    \label{fig:microsoft_like_flips}
\end{figure}

In Fig.~\ref{fig:microsoft_like_flips} and Tab.~\ref{tab:qwen-accuracy} we report accuracy results on various QA tasks. Note that flips (as reported in the main paper) are the more reliable (and less easily manipulated) metric than accuracy for QA tasks, as shown in~\cite{notaccuracy}. Fig.~\ref{fig:microsoft_like_flips} closely follows the analysis presented in prior work~\cite{notaccuracy}, further confirming the alignment of our findings with existing literature.
\begin{table*}[]
\centering
\small
\setlength{\tabcolsep}{4pt}
\caption{Accuracy (\%) on HellaSwag, PIQA, and MMLU for Qwen3 models with 3-bit and 4-bit quantization. Higher is better.}
\label{tab:qwen-accuracy}
\begin{tabular}{l l l*{8}{c}}
\toprule
& & & \multicolumn{4}{c}{\textbf{Qwen3-14B}} & \multicolumn{4}{c}{\textbf{Qwen3-32B}} \\
\cmidrule(lr){4-7} \cmidrule(lr){8-11}
& & Method
& \textit{HellaSwag} & \textit{PIQA} & \textit{MMLU} & \others{Avg.\,$\uparrow$}
& \textit{HellaSwag} & \textit{PIQA} & \textit{MMLU} & \others{Avg.\,$\uparrow$} \\
\midrule
& & Original (BF16) & \textit{60.95} & \textit{80.20} & \textit{78.83} & \others{73.33} & \textit{63.85} & \textit{80.96} & \textit{81.88} & \others{75.56} \\
\midrule

% Calibration-free
\multirow{10}{*}{\rotatebox[origin=c]{90}{\scriptsize\textsc{Calibration-free}}}
& \multirow{4}{*}{\rotatebox[origin=c]{90}{\scriptsize\textsc{3-bit}}}
   & RTN            & 56.99 & 77.80 & 75.01 & \others{69.93} & 46.98 & 71.82 & 78.53 & \others{65.78} \\
&  & Hadamard + RTN & 49.66 & 73.45 & 67.53 & \others{63.55} & 50.43 & 75.41 & 78.10 & \others{67.98} \\
&  & HQQ            & 55.50 & \textbf{77.91} & 72.92 & \others{68.78} & 59.87 & 77.75 & 78.17 & \others{71.93} \\
&  & \ours{\textbf{SINQ} \textit{(ours)}} & \ours{\textbf{58.03}} & \ours{77.20} & \ours{\textbf{75.82}} & \ours{\textbf{70.35}} & \ours{\textbf{60.65}} & \ours{\textbf{79.49}} & \ours{\textbf{78.89}} & \ours{\textbf{73.11}} \\
\cmidrule(lr){2-11}
& \multirow{6}{*}{\rotatebox[origin=c]{90}{\scriptsize\textsc{4-bit}}}
  & RTN              & 60.11 & 79.11 & \textbf{78.44} & \others{72.55} & 61.22 & 78.78 & 81.78 & \others{73.93} \\
& & BnB  (FP4)       & 59.41 & 79.38 & 77.62 & \others{72.14} & 56.97 & 77.91 & 81.20 & \others{72.03} \\
& & BnB (NF4)        & \textbf{60.47} & 79.71 & 78.23 & \others{72.80} & 63.12 & 79.98 & 81.60 & \others{74.90} \\
&  & Hadamard + RTN  & 58.45 & 78.67 & 76.58 & \others{71.23} & 62.90 & 78.94 & 80.99 & \others{74.28} \\
&  & HQQ             & 60.24 & 79.76 & 78.24 & \others{72.75} & 62.33 & 79.92 & \textbf{81.68} & \others{74.64} \\
&  & \ours{\textbf{SINQ} \textit{(ours)}} & \ours{60.05} & \ours{79.54} & \ours{78.00} & \ours{72.53} & \ours{\textbf{63.20}} & \ours{\textbf{80.85}} & \ours{81.63} & \ours{\textbf{75.23}} \\
&  & \ours{\textbf{SINQ} (NF4) \textit{(ours)}} & \ours{60.35} & \ours{\textbf{79.72}} & \ours{78.37} & \ours{\textbf{72.81}} & \ours{63.18} & \ours{80.52} & \ours{81.32} & \ours{75.00} \\
\midrule
% Calibrated
\multirow{7}{*}{\rotatebox[origin=c]{90}{\scriptsize\textsc{Calibrated}}}
& \multirow{3}{*}{\rotatebox[origin=c]{90}{\scriptsize\textsc{3-bit}}}
   & GPTQ             & \textbf{58.34} & 76.71 & 74.75 & \others{69.93} & 61.16 & 77.86 & 78.94 & \others{72.65} \\
&  & Hadamard + GPTQ  & 57.41 & \textbf{77.75} & 75.10 & \others{70.08} & 61.26 & 78.45 & 78.78 & \others{72.83} \\
&  & \ours{\textbf{A-SINQ} \textit{(ours)}} & \ours{58.16} & \ours{77.15} & \ours{\textbf{75.40}} & \ours{\textbf{70.24}} & \ours{\textbf{61.47}} & \ours{\textbf{79.22}} & \ours{\textbf{79.00}} & \ours{\textbf{73.23}} \\
\cmidrule(lr){2-11}
& \multirow{4}{*}{\rotatebox[origin=c]{90}{\scriptsize\textsc{4-bit}}}
   & GPTQ              & 60.55 & 79.43 & \textbf{78.11} & \others{72.70} & 63.22 & 80.20 & 81.36 & \others{74.93} \\
&  & Hadamard + GPTQ   & 60.27 & \textbf{79.60} & 77.85 & \others{72.57} & 63.01 & \textbf{81.01} & 80.98 & \others{75.00} \\
&  & AWQ               & 60.48 & 79.38 & 78.01 & \others{72.62} & \textbf{63.51} & 79.90 & 81.38 & \others{74.93} \\
&  & \ours{\textbf{A-SINQ} \textit{(ours)}} & \ours{\textbf{60.84}} & \ours{79.22} & \ours{78.07} & \ours{\textbf{72.71}} & \ours{63.43} & \ours{80.03} & \ours{\textbf{81.61}} & \ours{\textbf{75.02}} \\
\bottomrule
\bottomrule
\end{tabular}
\end{table*}

\subsection{Results on Phi Models}
\label{sec:phi}
In Tab.~\ref{tab:phi-ptq} we report quantization results on Phi family models. These findings further validate the effectiveness of SINQ also on this type of architecture. 

\begin{table*}[]
\centering
\small
\setlength{\tabcolsep}{3.6pt}
\caption{Weight-only PTQ on Phi models with 3-bit and 4-bit quantization, reporting perplexity and actual memory usage (GB). Lower is better for all metrics. In bold is the best result for a given setting.}
\label{tab:phi-ptq}
\begin{tabular}{lll*{9}{c}}
\toprule
& & 
& \multicolumn{3}{c}{\textbf{Phi-2} (3B)} 
& \multicolumn{3}{c}{\textbf{Phi-3} (4B)} 
& \multicolumn{3}{c}{\textbf{Phi-4} (15B)} \\
\cmidrule(lr){4-6} \cmidrule(lr){7-9} \cmidrule(lr){10-12}
& & Method
& Mem. & \others{\textit{Wiki2}\,$\downarrow$} & \others{\textit{C4}\,$\downarrow$}
& Mem. & \others{\textit{Wiki2}\,$\downarrow$} & \others{\textit{C4}\,$\downarrow$}
& Mem. & \others{\textit{Wiki2}\,$\downarrow$} & \others{\textit{C4}\,$\downarrow$} \\
\midrule

% Baseline (FP16)
& & Original (BF16) 
& 5.18 & \others{\textit{9.82}} & \others{\textit{13.83}}
& 7.11 & \others{\textit{6.01}} & \others{\textit{8.96}}
& 27.31 & \others{\textit{6.67}} & \others{\textit{11.13}} \\
\midrule

% Calibration-free
\multirow{6}{*}{\rotatebox[origin=c]{90}{\scriptsize\textsc{Calibration-free}}}
& \multirow{3}{*}{\rotatebox[origin=c]{90}{\scriptsize\textsc{3-bit}}}
   & RTN                  
   & 1.57 & \others{12.24} & \others{16.27}
   & 1.96 & \others{9.74} & \others{12.39}
   & 7.82 & \others{7.29} & \others{12.23} \\
& & HQQ                   
   & 1.57 & \others{11.37} & \others{15.69}
   & 1.96 & \others{11.60} & \others{16.42}
   & 7.82 & \others{7.41} & \others{15.60} \\
& & \ours{\textbf{SINQ} \textit{(ours)}} 
   & \ours{1.64} & \ours{\textbf{11.07}} & \ours{\textbf{15.23}}
   & \ours{1.99} & \ours{9.56} & \ours{12.14}
   & \ours{7.91} & \ours{\textbf{7.28}} & \ours{\textbf{12.19}} \\
\cmidrule(lr){2-12}

& \multirow{3}{*}{\rotatebox[origin=c]{90}{\scriptsize\textsc{4-bit}}}
 & RTN              
   & 1.81 & \others{10.30} & \others{14.40}
   & 2.28 & \others{6.95} & \others{9.71}
   & 9.18 & \others{6.64} & \others{11.38} \\
& & HQQ             
   & 1.81 & \others{10.09} & \others{14.23}
   & 2.28 & \others{6.85} & \others{9.70}
   & 9.18 & \others{6.80} & \others{14.84} \\
& & \ours{\textbf{SINQ} \textit{(ours)}} 
   & \ours{1.86} & \ours{\textbf{9.98}} & \ours{\textbf{14.09}}
   & \ours{2.29} & \ours{6.79} & \ours{9.68}
   & \ours{9.32} & \ours{\textbf{6.61}} & \ours{\textbf{11.32}} \\
\bottomrule
\bottomrule
\end{tabular}
\end{table*}

\subsection{Comparison to CrossQuant in W4A8 setting}
\label{sec:crossquant}
Here we compare to the CrossQuant method \cite{liu2024crossquant}. We separate these results from the main text, because CrossQuant uses a W4A8G128 setting, so that the values are not directly comparable to the main results (using W4A16G64) of the paper. See Tab.~\ref{tab:crossquant}. 

% In the activation-quantized setting, if utilizing low-precision matmuls in combination with SINQ, activation quantization needs to performed after input side scale is applied:
% \begin{equation}
%     \vec{y}_8 = \text{cast}_8(\vec{x}_{16} \odot \vec{t}_{16}) \cdot \text{cast}_8(\mathbf{W}_4, \vec{s}_{16}, \vec{z}_{16})
% \end{equation}
% We observe no noteworthy quality degradation from this activation quantization to 8 bits.

\begin{table}[H]
\centering
\small
\setlength{\tabcolsep}{10pt} % Slightly wider than 3.6pt since we have fewer columns here
\caption{Wikitext perplexity comparison to CrossQuant on Llama2 models (we use context length 2048, W4A8G128 to match reported CrossQuant results). Lower is better. In bold is the best result.}
\label{tab:crossquant}
\begin{tabular}{lcc}
\midrule
& \textbf{Llama2-7B} 
& \textbf{Llama2-13B} \\
\cmidrule(lr){2-2} \cmidrule(lr){3-3}
Method & \others{\textit{Wiki2}\,$\downarrow$} & \others{\textit{Wiki2}\,$\downarrow$} \\
\midrule
Original (BF-16)   & \others{5.47} & \others{4.88} \\
\toprule
CrossQuant & \others{5.79}          & \others{5.14} \\
\ours{ASINQ}      & \ours{\textbf{5.62}}          & \ours{\textbf{4.97}} \\
\bottomrule
\bottomrule
\end{tabular}
\end{table}

\subsection{Comparison to Code-Book-Based Methods}
Here we compare to two recent code-book-based methods by \cite{qtip} and \cite{hadamard}. Note that code-book-based methods are incompatible with activation quantization and require non-standard operations / kernels (may not work on NPU, TPU, mobile). See Tab.~\ref{tab:code-book}. 
\begin{table}[H]
\centering
\small
\setlength{\tabcolsep}{10pt}
\caption{Wikitext perplexity comparison to code-book-based models on Llama2 models (context length 4096). Note that code-book-based methods are incompatible with activation quantization and require non-standard operations (may not be NPU, TPU, mobile compatible).}
\label{tab:code-book}
\begin{tabular}{lcc}
\toprule
& \textbf{Llama2-7B} 
& \textbf{Llama2-13B} \\
\cmidrule(lr){2-2} \cmidrule(lr){3-3}
Method & \others{\textit{Wiki2}\,$\downarrow$} & \others{\textit{Wiki2}\,$\downarrow$} \\
\midrule
Baseline & \others{5.12} & \others{4.57} \\
\midrule
QTIP     & \others{\textbf{5.17}}          & \others{\textbf{4.62}} \\
QUIP\#   & \others{5.22}          & \others{4.65} \\
\ours{ASINQ}    & \ours{5.22}          & \ours{4.64} \\
\bottomrule
\bottomrule
\end{tabular}
\end{table}

\subsection{Further Comparison to HIGGS}
\label{sec:app:higgs}
%{\color{red} I would like a table here that compares HIGGS to \url{sinq_quantAux} to show we get better perplexity at lower memory (as HIGGS is the only(?) uncalibrated method that by default quantizes the auxiliaries). }
For a fairer comparison to the HIGGS method, in Tab.~\ref{tab:HIGGS} compare it to SINQ with quantized auxiliaries (to ensure more similar memory usage). 
\begin{table*}[]
\centering
\small
\setlength{\tabcolsep}{3.6pt}
\caption{Comparison to HIGGS method with quantized auxiliary variables to better match the HIGGS memory use.}
\label{tab:HIGGS}
\begin{tabular}{l@{} l l*{9}{r}}
\toprule
& & & \multicolumn{3}{c}{\textbf{Qwen3-1.7B}} & \multicolumn{3}{c}{\textbf{Qwen3-14B}} & \multicolumn{3}{c}{\textbf{Qwen3-32B}} \\
\cmidrule(lr){4-6} \cmidrule(lr){7-9} \cmidrule(lr){10-12}
& & Method 
& Mem. & \others{\textit{Wiki2}\,$\downarrow$} & \others{\textit{C4}\,$\downarrow$}
& Mem. & \others{\textit{Wiki2}\,$\downarrow$} & \others{\textit{C4}\,$\downarrow$}
& Mem. & \others{\textit{Wiki2}\,$\downarrow$} & \others{\textit{C4}\,$\downarrow$} \\
\midrule

% Baseline (FP16)
& & Original (BF16) & 3.44 & \others{\textit{16.67}} & \others{\textit{19.21}} & 29.54 & \others{\textit{8.64}} & \others{\textit{12.01}} & 65.52 & \others{\textit{7.60}} & \others{\textit{10.77}} \\
\midrule

& \multirow{3}{*}{\rotatebox[origin=c]{90}{\scriptsize\textsc{4-bit}}}
    & HIGGS (non-uniform)             & 1.51 & \others{23.98} & \others{25.27} & 10.28 & \others{9.13        } & \others{12.56} & 19.88 & \others{8.02} & \others{11.24} \\
& & \ours{\textbf{SINQ} (NF4) \textit{(ours)}} & \ours{1.42} & \ours{16.94} & \ours{\textbf{19.83}} & \ours{10.56} & \ours{\textbf{8.72}} & \ours{\textbf{12.13}} & \ours{20.73} & \ours{7.83} & \ours{\textbf{10.97}} \\
& & \ours{\textbf{SINQ} (NF4) \textit{(ours, q. aux.)}} &  \ours{\textbf{1.24}} & \ours{\textbf{16.92}}& \ours{19.84} & \ours{\textbf{10.19}} & {\textbf{\ours{8.72}}} & {\ours{\textbf{12.13}}} & \ours{\textbf{19.80}} & \ours{\textbf{7.82}} & \ours{10.98} \\
\bottomrule
\bottomrule
\end{tabular}
\end{table*}

% \subsection{Combination of SINQ and Hadamard Rotation}

% We find that combining Hadamard and SINQ does not further improve results. Intuitively, this is because both Hadamard rotation and SINQ aim to transform the space in which we quantize the matrix -- both succeed to some extent, with SINQ having an advantage, see Tab.~\ref{tab:sinq+had}.
% \input{tables/results_sinq+had}

\subsection{Additional Results on MoE Models}
\label{sec:app:moe}
In Tab.~\ref{tab:moe-ptq} we show some perplexity results on MoE models to underline the flexibility of our method. These results further demonstrate that SINQ is able to outperform state-of-the-art calibration-free methods for weight quantization.
\begin{table*}[]
\centering
\small
\setlength{\tabcolsep}{5pt}
\caption{Weight-only PTQ on \textbf{DeepSeek-V2-Lite} and \textbf{Qwen3-30B-A3B} MoE models with 3-bit and 4-bit quantization, reporting perplexity and actual memory usage (GB). Lower is better for all metrics. In bold is the best result for a given setting.}
\label{tab:moe-ptq}
\begin{tabular}{l l c c c c c c}
\toprule
& & \multicolumn{3}{c}{\textbf{DeepSeek-V2-Lite}} & \multicolumn{3}{c}{\textbf{Qwen3-30B-A3B}} \\
\cmidrule(lr){3-5} \cmidrule(lr){6-8}
Setting & Method & Mem. & \others{\textit{Wiki2}\,$\downarrow$} & \others{\textit{C4}\,$\downarrow$} 
                       & Mem. & \others{\textit{Wiki2}\,$\downarrow$} & \others{\textit{C4}\,$\downarrow$} \\
\midrule

Baseline & Original (BF16) & 32.55 & \others{\textit{6.31}} & \others{\textit{8.83}} 
                              & 61.06 & \others{\textit{8.70}} & \others{\textit{12.15}} \\
\midrule

Calibration-free (3-bit) & RTN & 9.12 & \others{7.94} & \others{10.98} 
                                      & 15.10 & \others{12.28} & \others{15.89} \\
                         & HQQ & 9.12 & \others{8.36} & \others{11.74} 
                                      & 15.10 & \others{10.52} & \others{14.39} \\
                         & \ours{\textbf{SINQ} \textit{(ours)}} 
                              & \ours{9.02} & \ours{\textbf{7.45}} & \ours{\textbf{10.32}} 
                              & \ours{15.13} & \ours{\textbf{10.19}} & \ours{\textbf{13.62}} \\
\midrule

Calibration-free (4-bit) & RTN & 10.63 & \others{6.59} & \others{9.19} 
                                      & 18.07 & \others{9.04} & \others{12.64} \\
                         & BnB      &  10.63 & \others{6.82} & \others{9.49} 
                                      & 18.08 & \others{9.68} & \others{12.93} \\
                         & HQQ & 10.85 & \others{6.61} & \others{9.18} 
                                      & 18.07 & \others{9.14} & \others{12.64} \\
                         & \ours{\textbf{SINQ} \textit{(ours)}} 
                              & \ours{10.50} & \ours{\textbf{6.49}} & \ours{\textbf{9.07}} 
                              & \ours{18.13} & \ours{\textbf{9.02}} & \ours{\textbf{12.41}} \\
% \midrule

% Calibrated (3-bit) & \ours{\textbf{A-SINQ} \textit{(ours)}} 
%                               & \ours{--} & \ours{--} & \ours{--} 
%                               & \ours{--} & \ours{--} & \ours{--} \\
% \midrule

% Calibrated (4-bit) & AWQ & -- & \others{--} & \others{--} 
%                                & -- & \others{--} & \others{--} \\
%                    & \ours{\textbf{A-SINQ} \textit{(ours)}} 
%                                & \ours{--} & \ours{\textbf{--}} & \ours{\textbf{--}} 
%                                & \ours{--} & \ours{\textbf{--}} & \ours{\textbf{--}} \\
\bottomrule
\bottomrule
\end{tabular}
\end{table*}

% \begin{algorithm}
% \caption{Round to Nearest Asymmetric Quantization (Unsigned)}
% \label{alg:asymmetric_quant_unsigned}
% \begin{algorithmic}[2]
% \Require $\mathbf{X} \in \mathbb{R}^{m \times n}$, bits
% \Ensure $\mathbf{Q} \in \mathbb{Z}^{m \times n}$, $\bm{\Delta}$, $\mathbf{z}$

% \State $q_{\text{min}} \gets 0$, $q_{\text{max}} \gets 2^{\text{bits}} - 1$
% \State $\bm{\Delta} \gets \mathbf{0}^n$, $\mathbf{z} \gets \mathbf{0}^n$, $\mathbf{Q} \gets \mathbf{0}^{m \times n}$

% \For{$j \gets 1$ to $n$}
%     \State $min_j \gets \min(\mathbf{X}[:, j])$, $max_j \gets \max(\mathbf{X}[:, j])$
%     \State $\bm{\Delta}[j] \gets (max_j - min_j) / (q_{\text{max}} - q_{\text{min}})$
%     \State $\mathbf{z}[j] \gets \left\lfloor -min_j / \bm{\Delta}[j] + q_{\text{min}} + 0.5 \right\rfloor$
%     \State $\mathbf{z}[j] \gets \min(\max(\mathbf{z}[j], q_{\text{min}}), q_{\text{max}})$
    
%     \For{$i \gets 1$ to $m$}
%         \State $q_{\text{float}} \gets \mathbf{X}[i, j] / \bm{\Delta}[j] - \mathbf{z}[j]$
%         \State $\mathbf{Q}[i, j] \gets \left\lfloor q_{\text{float}} + 0.5 \right\rfloor$
%         \State $\mathbf{Q}[i, j] \gets \min(\max(\mathbf{Q}[i, j], q_{\text{min}}), q_{\text{max}})$
%     \EndFor
% \EndFor

% \State \Return $\mathbf{Q}, \bm{\Delta}, \mathbf{z}$
% \end{algorithmic}
% \end{algorithm}


\begin{thebibliography}{29}
\providecommand{\natexlab}[1]{#1}
\providecommand{\url}[1]{\texttt{#1}}
\expandafter\ifx\csname urlstyle\endcsname\relax
  \providecommand{\doi}[1]{doi: #1}\else
  \providecommand{\doi}{doi: \begingroup \urlstyle{rm}\Url}\fi

\bibitem[Akhondzadeh et~al.()Akhondzadeh, Bojchevski, Eleftheriou, and Dazzi]{kurtail}
Akhondzadeh, M.~S., Bojchevski, A., Eleftheriou, E., and Dazzi, M.
\newblock Kurtail: Kurtosis-based llm quantization.
\newblock In \emph{Sparsity in LLMs (SLLM): Deep Dive into Mixture of Experts, Quantization, Hardware, and Inference}.

\bibitem[Ashkboos et~al.(2024)Ashkboos, Mohtashami, Croci, Li, Cameron, Jaggi, Alistarh, Hoefler, and Hensman]{quarot}
Ashkboos, S., Mohtashami, A., Croci, M.~L., Li, B., Cameron, P., Jaggi, M., Alistarh, D., Hoefler, T., and Hensman, J.
\newblock Quarot: Outlier-free 4-bit inference in rotated llms.
\newblock \emph{Advances in Neural Information Processing Systems}, 37:\penalty0 100213--100240, 2024.

\bibitem[Badri \& Shaji(2023)Badri and Shaji]{HQQ}
Badri, H. and Shaji, A.
\newblock Half-quadratic quantization of large machine learning models, November 2023.
\newblock URL \url{https://mobiusml.github.io/hqq_blog/}.

\bibitem[{Badri et al.}(2024)]{gemlite}
{Badri et al.}
\newblock Gemlite: Triton kernels for efficient low-bit matrix multiplication, 2024.
\newblock URL \url{https://github.com/dropbox/gemlite}.

\bibitem[DeepSeek-AI(2024)]{DS-V2}
DeepSeek-AI.
\newblock Deepseek-v2: A strong, economical, and efficient mixture-of-experts language model, 2024.

\bibitem[Dettmers et~al.(2022)Dettmers, Lewis, Belkada, and Zettlemoyer]{llmint8}
Dettmers, T., Lewis, M., Belkada, Y., and Zettlemoyer, L.
\newblock Gpt3.int8 (): 8-bit matrix multiplication for transformers at scale.
\newblock \emph{Advances in neural information processing systems}, 35:\penalty0 30318--30332, 2022.

\bibitem[Dettmers et~al.(2023)Dettmers, Pagnoni, Holtzman, and Zettlemoyer]{NF4}
Dettmers, T., Pagnoni, A., Holtzman, A., and Zettlemoyer, L.
\newblock Qlora: Efficient finetuning of quantized llms.
\newblock \emph{Advances in neural information processing systems}, 36:\penalty0 10088--10115, 2023.

\bibitem[Dutta et~al.(2024)Dutta, Krishnan, Kwatra, and Ramjee]{notaccuracy}
Dutta, A., Krishnan, S., Kwatra, N., and Ramjee, R.
\newblock Accuracy is not all you need.
\newblock \emph{Advances in Neural Information Processing Systems}, 37:\penalty0 124347--124390, 2024.

\bibitem[Elhoushi \& Johnson(2025)Elhoushi and Johnson]{any4}
Elhoushi, M. and Johnson, J.
\newblock any4: Learned 4-bit numeric representation for llms.
\newblock \emph{arXiv preprint arXiv:2507.04610}, 2025.

\bibitem[Fedus et~al.(2022)Fedus, Zoph, and Shazeer]{MoE}
Fedus, W., Zoph, B., and Shazeer, N.
\newblock Switch transformers: Scaling to trillion parameter models with simple and efficient sparsity.
\newblock \emph{Journal of Machine Learning Research}, 23\penalty0 (120):\penalty0 1--39, 2022.

\bibitem[Frantar et~al.(2022)Frantar, Ashkboos, Hoefler, and Alistarh]{GPTQ}
Frantar, E., Ashkboos, S., Hoefler, T., and Alistarh, D.
\newblock Gptq: Accurate post-training quantization for generative pre-trained transformers.
\newblock \emph{arXiv preprint arXiv:2210.17323}, 2022.

\bibitem[Kingma \& Ba(2014)Kingma and Ba]{adam}
Kingma, D.~P. and Ba, J.
\newblock Adam: A method for stochastic optimization, 2014.
\newblock URL \url{https://arxiv.org/abs/1412.6980}.

\bibitem[Lin et~al.(2024{\natexlab{a}})Lin, Xu, Wu, Cui, Zhang, Mou, Song, Sun, and Wei]{duquant}
Lin, H., Xu, H., Wu, Y., Cui, J., Zhang, Y., Mou, L., Song, L., Sun, Z., and Wei, Y.
\newblock Duquant: Distributing outliers via dual transformation makes stronger quantized llms.
\newblock \emph{Advances in Neural Information Processing Systems}, 37:\penalty0 87766--87800, 2024{\natexlab{a}}.

\bibitem[Lin et~al.(2024{\natexlab{b}})Lin, Tang, Tang, Yang, Chen, Wang, Xiao, Dang, Gan, and Han]{AWQ}
Lin, J., Tang, J., Tang, H., Yang, S., Chen, W.-M., Wang, W.-C., Xiao, G., Dang, X., Gan, C., and Han, S.
\newblock Awq: Activation-aware weight quantization for on-device llm compression and acceleration.
\newblock \emph{Proceedings of machine learning and systems}, 6:\penalty0 87--100, 2024{\natexlab{b}}.

\bibitem[Liu et~al.(2024{\natexlab{a}})Liu, Feng, Xue, Wang, Wu, Lu, Zhao, Deng, Zhang, Ruan, et~al.]{DSV3}
Liu, A., Feng, B., Xue, B., Wang, B., Wu, B., Lu, C., Zhao, C., Deng, C., Zhang, C., Ruan, C., et~al.
\newblock Deepseek-v3 technical report.
\newblock \emph{arXiv preprint arXiv:2412.19437}, 2024{\natexlab{a}}.

\bibitem[Liu et~al.(2024{\natexlab{b}})Liu, Ma, Zhang, and Wang]{liu2024crossquant}
Liu, W., Ma, X., Zhang, P., and Wang, Y.
\newblock Crossquant: A post-training quantization method with smaller quantization kernel for precise large language model compression.
\newblock \emph{arXiv preprint arXiv:2410.07505}, 2024{\natexlab{b}}.

\bibitem[Liu et~al.()Liu, Zhao, Fedorov, Soran, Choudhary, Krishnamoorthi, Chandra, Tian, and Blankevoort]{spinquant}
Liu, Z., Zhao, C., Fedorov, I., Soran, B., Choudhary, D., Krishnamoorthi, R., Chandra, V., Tian, Y., and Blankevoort, T.
\newblock Spinquant: Llm quantization with learned rotations.
\newblock In \emph{The Thirteenth International Conference on Learning Representations}.

\bibitem[Ma et~al.()Ma, Li, Zheng, Ling, Xiao, Wang, Wen, Chao, and Ji]{affinequant}
Ma, Y., Li, H., Zheng, X., Ling, F., Xiao, X., Wang, R., Wen, S., Chao, F., and Ji, R.
\newblock Affinequant: Affine transformation quantization for large language models.
\newblock In \emph{The Twelfth International Conference on Learning Representations}.

\bibitem[Malinovskii et~al.(2025)Malinovskii, Panferov, Ilin, Guo, Richt{\'a}rik, and Alistarh]{HIGGS}
Malinovskii, V., Panferov, A., Ilin, I., Guo, H., Richt{\'a}rik, P., and Alistarh, D.
\newblock Higgs: Pushing the limits of large language model quantization via the linearity theorem.
\newblock In \emph{Proceedings of the 2025 Conference of the Nations of the Americas Chapter of the Association for Computational Linguistics: Human Language Technologies (Volume 1: Long Papers)}, pp.\  10857--10886, 2025.

\bibitem[Raffel et~al.(2020)Raffel, Shazeer, Roberts, Lee, Narang, Matena, Zhou, Li, and Liu]{c4}
Raffel, C., Shazeer, N., Roberts, A., Lee, K., Narang, S., Matena, M., Zhou, Y., Li, W., and Liu, P.~J.
\newblock Exploring the limits of transfer learning with a unified text-to-text transformer.
\newblock \emph{Journal of machine learning research}, 21\penalty0 (140):\penalty0 1--67, 2020.

\bibitem[Shao et~al.()Shao, Chen, Zhang, Xu, Zhao, Li, Zhang, Gao, Qiao, and Luo]{omniquant}
Shao, W., Chen, M., Zhang, Z., Xu, P., Zhao, L., Li, Z., Zhang, K., Gao, P., Qiao, Y., and Luo, P.
\newblock Omniquant: Omnidirectionally calibrated quantization for large language models.
\newblock In \emph{The Twelfth International Conference on Learning Representations}.

\bibitem[Sun et~al.()Sun, Liu, Bai, Bao, Zhao, Li, Yu, Hou, Yuan, Jiang, et~al.]{flatquant}
Sun, Y., Liu, R., Bai, H., Bao, H., Zhao, K., Li, Y., Yu, X., Hou, L., Yuan, C., Jiang, X., et~al.
\newblock Flatquant: Flatness matters for llm quantization.
\newblock In \emph{Forty-second International Conference on Machine Learning}.

\bibitem[Tseng et~al.(2024{\natexlab{a}})Tseng, Chee, Sun, Kuleshov, and De~Sa]{hadamard}
Tseng, A., Chee, J., Sun, Q., Kuleshov, V., and De~Sa, C.
\newblock Quip $\# $: Even better llm quantization with hadamard incoherence and lattice codebooks.
\newblock In \emph{International Conference on Machine Learning}, pp.\  48630--48656. PMLR, 2024{\natexlab{a}}.

\bibitem[Tseng et~al.(2024{\natexlab{b}})Tseng, Sun, Hou, and De~Sa]{qtip}
Tseng, A., Sun, Q., Hou, D., and De~Sa, C.~M.
\newblock Qtip: Quantization with trellises and incoherence processing.
\newblock \emph{Advances in Neural Information Processing Systems}, 37:\penalty0 59597--59620, 2024{\natexlab{b}}.

\bibitem[Xiao et~al.(2023)Xiao, Lin, Seznec, Wu, Demouth, and Han]{smoothquant}
Xiao, G., Lin, J., Seznec, M., Wu, H., Demouth, J., and Han, S.
\newblock Smoothquant: Accurate and efficient post-training quantization for large language models.
\newblock In \emph{International conference on machine learning}, pp.\  38087--38099. PMLR, 2023.

\bibitem[Yang et~al.(2025)Yang, Li, Yang, Zhang, Hui, Zheng, Yu, Gao, Huang, Lv, et~al.]{qwen3}
Yang, A., Li, A., Yang, B., Zhang, B., Hui, B., Zheng, B., Yu, B., Gao, C., Huang, C., Lv, C., et~al.
\newblock Qwen3 technical report.
\newblock \emph{arXiv preprint arXiv:2505.09388}, 2025.

\bibitem[Ye et~al.(2025)Ye, Xiao, Mi, and Liu]{aime}
Ye, Y., Xiao, Y., Mi, T., and Liu, P.
\newblock Aime-preview: A rigorous and immediate evaluation framework for advanced mathematical reasoning.
\newblock \url{https://github.com/GAIR-NLP/AIME-Preview}, 2025.
\newblock GitHub repository.

\bibitem[Zheng et~al.(2024)Zheng, Yin, Xie, Sun, Huang, Yu, Cao, Kozyrakis, Stoica, Gonzalez, et~al.]{sglang}
Zheng, L., Yin, L., Xie, Z., Sun, C.~L., Huang, J., Yu, C.~H., Cao, S., Kozyrakis, C., Stoica, I., Gonzalez, J.~E., et~al.
\newblock Sglang: Efficient execution of structured language model programs.
\newblock \emph{Advances in neural information processing systems}, 37:\penalty0 62557--62583, 2024.

\bibitem[Zheng et~al.(2025)Zheng, Li, Chu, Feng, Ma, Luo, Guo, Qin, Magno, and Liu]{qwenquant}
Zheng, X., Li, Y., Chu, H., Feng, Y., Ma, X., Luo, J., Guo, J., Qin, H., Magno, M., and Liu, X.
\newblock An empirical study of qwen3 quantization.
\newblock \emph{arXiv preprint arXiv:2505.02214}, 2025.

\end{thebibliography}
\end{document}